\PassOptionsToPackage{dvipsnames}{xcolor}
\documentclass[review]{elsarticle}
\makeatletter
\def\ps@pprintTitle{%
 \let\@oddhead\@empty
 \let\@evenhead\@empty
 \def\@oddfoot{}%
 \let\@evenfoot\@oddfoot}
\makeatother
\usepackage{tikz}
\usepackage{comment}
\usepackage{amsmath,amssymb} % define this before the line numbering.
\usepackage{color}
\usepackage{epsfig}
\usepackage[inline]{enumitem}
\usepackage[caption=false]{subfig}
\usepackage{soul}
\usepackage[utf8]{inputenc} % allow utf-8 input
\usepackage[T1]{fontenc}    % use 8-bit T1 fonts
\usepackage{booktabs}       % professional-quality tables
\usepackage{amsfonts}       % blackboard math symbols
\usepackage{nicefrac}       % compact symbols for 1/2, etc.
\usepackage{microtype}      % microtypography
\usepackage{algorithmic}
\usepackage{algorithm}
\usepackage{numprint}
\usepackage{siunitx}
\usepackage{etoolbox}% <-- for bold fonts
\newcommand{\ubold}{\fontseries{b}\selectfont}% <-- for bold fonts
\robustify\ubold% <-- for bold fonts
\usepackage{bbold}
\usepackage{wrapfig}
\usepackage{bm}
\usepackage[caption=false]{subfig}
\usepackage{pifont}% http://ctan.org/pkg/pifont
\usepackage{lineno,hyperref}
\usepackage{array,multirow,graphicx}

\usepackage[dvipsnames]{xcolor}
\makeatletter
\newcommand\footnoteref[1]{\protected@xdef\@thefnmark{\ref{#1}}\@footnotemark}
\makeatother

\modulolinenumbers[5]

\journal{Journal of Pattern Recognition}

\begin{document}

\pagenumbering{gobble}

\let\WriteBookmarks\relax
\def\floatpagepagefraction{1}
\def\textpagefraction{.001}
\pagenumbering{arabic}

\newcommand{\ttinneriter}{\texttt{inner\_iters}}
\newcommand{\ttit}{\texttt{iter}}
\newcommand{\ttmaxiter}{\texttt{max\_iter}}
\newcommand{\ttbestScore}{\texttt{best\_score}}
\newcommand{\ttbestR}{\texttt{best\_rotation}}
\newcommand{\ttbestT}{\texttt{best\_translation}}

\newcommand{\cL}{\mathcal{L}}
\newcommand{\cM}{\mathcal{M}}
\newcommand{\cN}{\mathcal{N}}
\newcommand{\cI}{\mathcal{I}}
\newcommand{\cS}{\mathcal{S}}
\newcommand{\cD}{\mathcal{D}}
\newcommand{\cP}{\mathcal{P}}
\newcommand{\cQ}{\mathcal{Q}}
\newcommand{\cO}{\mathcal{O}}
\newcommand{\cT}{\mathcal{T}}
\newcommand{\cad}{\mathcal{d}}
\newcommand{\cX}{\mathcal{X}}
\newcommand{\cXh}{\hat{\mathcal{X}}}
\newcommand{\cC}{\mathcal{C}}
\newcommand{\cF}{\mathcal{F}}

\newcommand{\be}{\mathbf{e}}
\newcommand{\br}{\mathbf{r}}
\newcommand{\bx}{\mathbf{x}}
\newcommand{\bxh}{\hat{\mathbf{x}}}
\newcommand{\bX}{\mathbf{X}}
\newcommand{\bY}{\mathbf{Y}}
\newcommand{\bZero}{\mathbf{0}}
\newcommand{\hbX}{\hat{\mathbf{X}}}
\newcommand{\bS}{\mathbf{S}}
\newcommand{\bs}{\mathbf{s}}
\newcommand{\bp}{\mathbf{p}}
\newcommand{\bq}{\mathbf{q}}
\newcommand{\bD}{\mathbf{D}}
\newcommand{\bd}{\mathbf{d}}
\newcommand{\bA}{\mathbf{A}}
\newcommand{\bR}{\mathbf{R}}
\newcommand{\bt}{\mathbf{t}}
\newcommand{\bH}{\mathbf{H}}
\newcommand{\bh}{\mathbf{h}}
\newcommand{\by}{\mathbf{y}}
\newcommand{\bz}{\mathbf{z}}
\newcommand{\bu}{\mathbf{u}}
\newcommand{\ba}{\mathbf{a}}
\newcommand{\bg}{\mathbf{g}}
\newcommand{\bo}{\mathbf{o}}
\newcommand{\bl}{\mathbf{l}}
\newcommand{\bOnes}{\mathbf{1}}
\newcommand{\bF}{\mathbf{F}}
\newcommand{\bK}{\mathbf{K}}
\newcommand{\bI}{\mathbf{I}}
\newcommand{\tdf}{\tilde{f}}
\newcommand{\tdh}{\tilde{h}}

\newcommand{\bbR}{\mathbb{R}}
\newcommand{\bbE}{\mathbb{E}}
\newcommand{\bbD}{\mathbb{D}}
\newcommand{\bbF}{\mathbb{F}}
\newcommand{\bbFh}{\hat{\mathbb{F}}}
\newcommand{\bmu}{\boldsymbol{\mu}}
\newcommand{\bhr}{\hat{\mathbf{r}}}
\newcommand{\bJ}{\mathbf{J}}
\newcommand{\Nsample}{$N_\text{sample}$}

\newcommand{\kernel}{\psi}

\newcommand{\residual}{\mathbf{r}}
\newcommand{\btheta}{\boldsymbol{\theta}}

\newcommand{\cV}{\mathcal{V}}
\newcommand{\cE}{\mathcal{E}}
\newcommand{\cB}{\mathcal{B}}
\newcommand{\cG}{\mathcal{G}}
\newcommand{\etal}{\textit{et al.}}
\newcommand{\xmark}{\ding{55}}%
\newcommand{\cmark}{\ding{51}}%

\begin{frontmatter}

\title{Kinematics Modeling Network for Video-based Human Pose Estimation}

\author[1]{Yonghao Dang}
\ead{dyh2018@bupt.edu.cn}
\author[1]{Jianqin Yin \corref{cor1}}
\cortext[cor1]{Corresponding author}
\author[1]{Shaojie Zhang}
\author[2,3,4]{Jiping Liu}
\author[5]{Yanzhu Hu}

\address[1]{School of Artificial Intelligence, Beijing University of Posts and Telecommunications, Beijing, China}
% \address[2]{Queen Mary School Hainan, Beijing University of Posts and Telecommunications}
\address[2]{Chinese Academy of Surveying and Mapping, Beijing, China}
\address[3]{Faculty of Geosciences and Environmental Engineering, Southwest Jiaotong University, Chengdu, China}
\address[4]{School of Resource and Environmental Sciences, Wuhan University, Wuhan, China}
\address[5]{School of Modern Post (School of Automation), Beijing University of Posts and Telecommunications, Beijing, China}

% \maketitle

%%%%%%%%% ABSTRACT
\begin{abstract}

Estimating human poses from videos is critical in human-computer interaction. Joints cooperate rather than move independently during human movement. There are both spatial and temporal correlations between joints. Despite the positive results of previous approaches, most of them focus on modeling the spatial correlation between joints while only straightforwardly integrating features along the temporal dimension, which ignores the temporal correlation between joints. In this work, we propose a plug-and-play kinematics modeling module (KMM) to explicitly model temporal correlations between joints across different frames by calculating their temporal similarity. In this way, KMM can capture motion cues of the current joint relative to all joints in different time. Besides, we formulate video-based human pose estimation as a Markov Decision Process and design a novel kinematics modeling network (KIMNet) to simulate the Markov Chain, allowing KIMNet to locate joints recursively. Our approach achieves state-of-the-art results on two challenging benchmarks. In particular, KIMNet shows robustness to the occlusion. Code will be released at \href{https://github.com/YHDang/KIMNet}{https://github.com/YHDang/KIMNet}.
\end{abstract}
\end{frontmatter}

%%%%%%%%% BODY TEXT
\section{Introduction}
\label{sec:intro}
Human pose estimation (HPE) plays a fundamental role in pattern recognition. Pose sequences exhibit resilience against superficial visual variations, such as the background, clothing texture, and illumination conditions \cite{UPS}, enabling pose sequences to offer pure action representations for other computer vision tasks, including action recognition \cite{RPAN}, person re-identification \cite{PersonRID}, and human parsing \cite{NRPose}. Enhancing the performance of HPE is significant for advancing the field of pattern recognition. According to data type, HPE can be roughly divided into image-based and video-based HPE. Unlike images, videos contain appearance features for each frame and encapsulate valuable temporal information. As a result, effectively modeling this temporal information becomes crucial for accurately estimating poses from videos.

 \begin{figure}[htbp]
  \centering
  \includegraphics[scale=0.35]{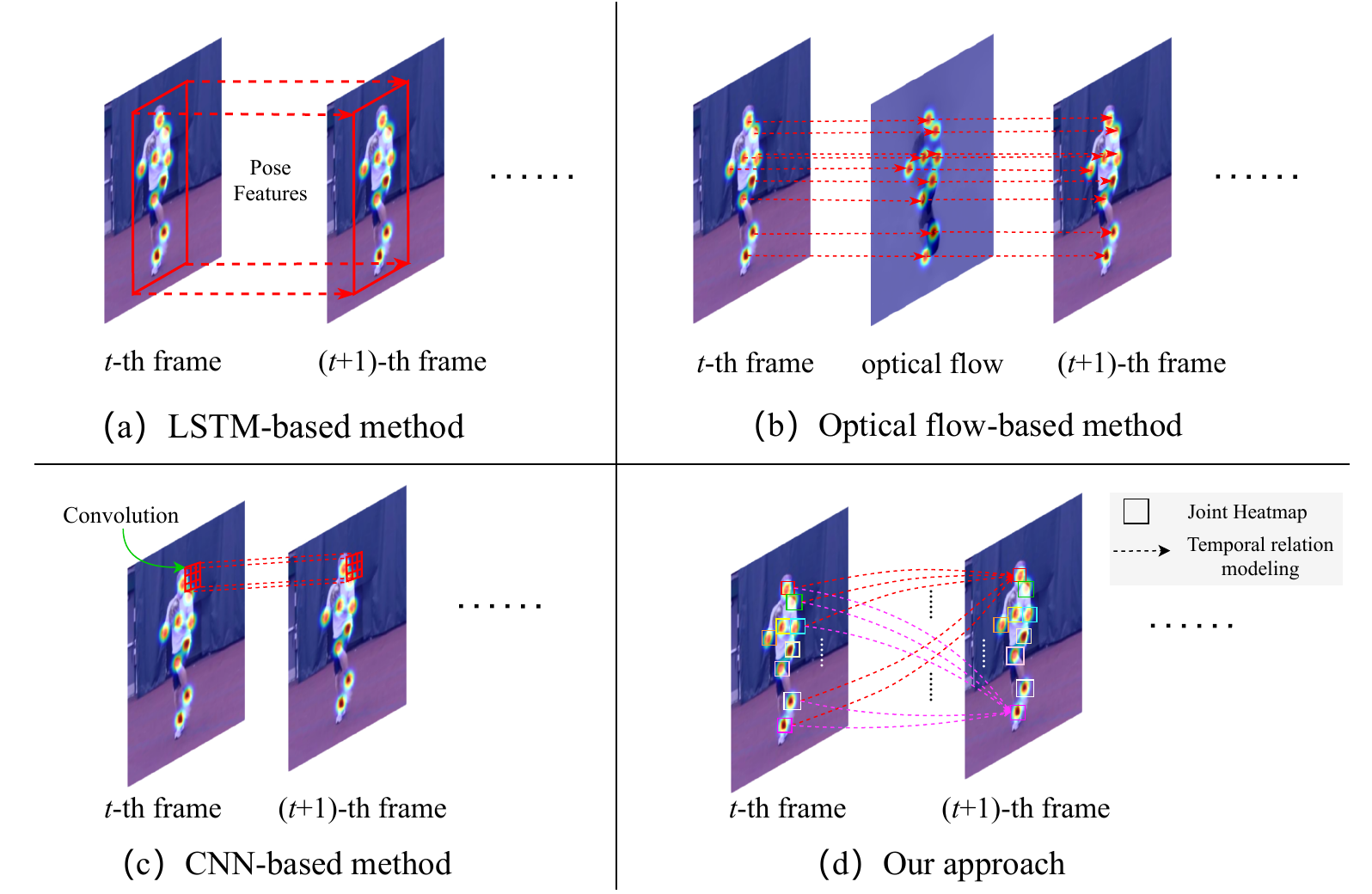}\\
  \caption{Comparisons between our approach and other existing methods. (a) LSTM-based methods. (b) Optical flow-based approaches. (c) CNN-based methods. (d) The proposed KIMNet.}\label{fig:fig1_comp}
 \end{figure}

 Due to its remarkable progress in the computer vision community, deep learning approaches have been commonly used to model temporal information for video-based HPE. As shown in Figure\ref{fig:fig1_comp}, LSTM-based methods \cite{LSTMPose,UniPose} aggregate vanilla sequential features of poses in adjacent frames. Optical flow-based \cite{FlowConv,ThinSlicing} approaches align the same joints in different frames through the joints' optical flow information. CNN-based \cite{3DHRNet,TCE,DKD,RPSTN} methods model the motion information of poses by integrating local features within the convolutional kernel along the temporal dimension. Despite the positive results shown in previous methods, there is a commonly essential but neglected factor among these approaches. Most methods model the motion information of the poses in the temporal dimension but ignore the temporal correlation between different joints.
 
 The joints of the human associate with each other rather than moving independently during human movement. In other words, there is simultaneous cooperation among all joints to maintain coordination of actions \cite{CA}. The cooperation of joints can be viewed as the spatial and temporal dependency between joints. The spatial dependency between joints can be regarded as the pose structural information that is represented by the spatial correlations between joints \cite{RPSTN}. Meanwhile, the temporal dependency between joints can be seen as the relative motion between different joints. For a clip of video, the joints in frame $t+1$ move relative to all joints in frame $t$. This relative motion can be expressed as the temporal correlation between joints to some content. Based on the temporal correlation, the model can infer the positions of joints in the current frame via the information about joints in the previous frames. The model can infer its position for the occluded joint according to the information of joints related to the occluded joint in other frames. Therefore, the temporal dependency between joints is significant for locating joints. However, most approaches pay much attention to modeling the pose structural information, while ignoring the temporal correlation between joints, which limits the model's performance for estimating poses from videos.
 
 %There have been many studies \cite{RPSTN,PGCN,ORGM} to model the spatial dependency between joints. However, the spatial dependency usually depends on the spatial representations of joints in a single frame. The spatial dependency between joints is usually unreliable when the spatial information of joints is incomplete, such as the occlusion and motion blur. Unlike spatial dependency, the temporal dependency between joints is not limited to the information within a single frame but provides the model more extra pose information from historical frames, alleviating the problem of incomplete spatial information of joints in a single frame. For example, the occluded joints' spatial information is fragmentary in a single frame. If the temporal dependency between joints is captured, the model can infer the positions of occluded joints according to the information of joints related to the occluded joints in other frames. Therefore, the temporal dependency between joints is significant for locating joints.

 In this paper, we propose a \textbf{KI}nematics \textbf{M}odeling based \textbf{Net}work (KIMNet) to locate joints by explicitly modeling the temporal dependency between joints across different frames. Specifically, we present a novel and plug-and-play kinematic modeling module (KMM) based on the attention mechanism to explicitly explore the temporal correlation between joints by modeling their temporal similarity. In this way, the KMM learns the motion cue of each joint relative to all joints in the previous frame. By combining motion cues and the historically positional information of poses, KMM can preliminarily predict the initial positions of joints from the current frame in advance. Based on this, we formulate the video-based human pose estimation as a Markov decision process. The proposed KIMNet recursively estimates human poses from each frame. Besides, thanks to the temporal dependency, the proposed KIMNet locates the current joint by aggregating the information of other joints related to the current joint instead of locating it only based on its own features.

 Contributions of this work are summarized as follows.
 \begin{itemize}
   \item We propose a plug-and-play kinematics modeling module (KMM) based on the attention mechanism to explicitly model the temporal correlation between joints across different frames. The KMM can predict the initial joints' positions in advance by aggregating joints' motion information and historical positions.
   
   \item We formulate video-based human pose estimation as a Markov decision process and present a \textbf{KI}nematics \textbf{M}odeling based \textbf{Net}work (KIMNet) to simulate it. With the guidance of the temporal correlation, KIMNet locates the current joint by integrating the information of other joints from other frames rather than only depending on the current joint's information, which improves the model's robustness against the occlusion.

   \item The proposed KIMNet achieves new state-of-the-art performance among methods on the challenging Penn Action and Sub-JHMDB datasets. Furthermore, Experimental results demonstrate that the KMM is compatible with existing pose estimation frameworks.
 \end{itemize}

\section{Related Works}
\subsection{Video-based Human Pose Estimation}
 Human pose estimation has always been one of the hot issues in computer vision tasks. With the improvement of deep learning, image-based pose estimation \cite{Hourglass,SimpleBaseline,HRNet,PGCN} has made significant progress. However, due to the lack of temporal modeling, image-based HPE approaches are difficult to maintain superior performance in video-based HPE.
 
 There has been a lot of work trying to explore temporal information of poses. The optical flow \cite{FlowConv,ThinSlicing} was first used to align joints across different frames. These methods have achieved attractive results while suffering from high computational cost. \cite{LSTMPose,UniPose} applied LSTM to extract temporal features. LSTM shows superior performance on temporal modeling, but the above methods are constrained to model the sequential correlation based on image features simply. Recently, CNN \cite{3DHRNet,DKD,KFPN} is commonly used to extract spatio-temporal representation of poses. The spatio-temporal features are usually limited by the receptive field of convolutional kernels for these CNN-based approaches. \cite{DiffPose} proposed a diffusion architecture to aggregate visual evidence across frames. \cite{TFAM} leveraged a hierarchical framework to capture coarse-to-fine deformations of poses across frames. Unlike these methods that focus on enhancing the temporal features of poses, the proposed method aims to capture the motion of poses by explicitly modeling the temporal correlation between joints across frames.
 
\subsection{Relation Modeling in Human Pose Estimation}

 The effectiveness of relation modeling has been demonstrated in many fields, such as image re-ranking \cite{ImageReranking}, image \cite{VLAAD} and action recognition \cite{UPS}. Yu et al. \cite{ImageReranking} proposed a multimodal hypergraph learning-based sparse coding to explore the complementarity of different features. Zhang et al.\cite{VLAAD} presented a vector of locally and adaptively aggregated descriptors to improve image feature representation. Compared with the image, the human body is more structural because of the apparent connection between joints. To model the structural information of the human body, Bin et al. \cite{PGCN} utilized a graphic convolution to model the physical connection between joints. \cite{ORGM} designed a graph structure network (GSN) to locate invisible joints by aggregating other joints' information. Jiang et al. \cite{PGA-Net} pointed out that multi-scale features of poses are significant for locating joints and proposed a pyramid gating network to capture semantic features of poses. Yang et al.\cite{DGNet} used pairwise attention to explore the association between joints. Dang et al. \cite{RPSTN} introduced a relation modeling module based on an attention mechanism to explore the structural information of poses. Yang et al. \cite{DGNet} used a self-attention mechanism to explore the relationship between different instances' joints. Similarly, \cite{GRRPT,UformPose} used a transformer to model the spatial correlations between pose tokens. Most methods mentioned above focus on spatial correlation modeling. Gai et al. \cite{SLT-Pose} proposed an SLT-Pose to strengthen the interaction of poses between the target frame and the local sequence through the cross-attention mechanism. Different from \cite{SLT-Pose} that explore the pose-level interaction across frames, the proposed KIMNet is able to model the joint-level temporal correlations.

\subsection{Temporal Correlation Modeling in Video Analysis}
 Temporal correlations have been widely used in various tasks. Wu et al. \cite{3DPersonReID} used the attention mechanism for integrating spatiotemporal features in a person re-identification task. Wang et al. introduced pixel-wise contrastive algorithm \cite{PixelCorrSS} and co-attention classifier \cite{WSSS} to associate the salient objects for the semantic segmentation task. Furthermore, Wang et al. proposed COSNet \cite{COSNet} with the group co-attention mechanism \cite{COSNetZeroShot} to predict object masks and associate them across multiple frames for the video-based object segmentation task. Similar to video-based object segmentation, object tracking also needs to match the same target in different frames. Wu et al. \cite{TraDes} presented an online TraDeS tracker that assigns the same target in multiple frames by calculating a pixel-wise similarity between adjacent frames. The above approaches usually aim at associating objects from different frames by modeling their correlations at the pixel level. For the human pose estimation, it is worth focusing on the correlation between the whole joints (represented by feature maps or heatmaps). The pixel-level correlation between joints' features will destroy the integrity of joints to some content. To solve this issue, KMM is proposed to model the temporal correlations between joints across frames.

\section{Methodology}

 \subsection{Problem Statement}
 \label{sec:section_A}
 We formulate the video-based human pose estimation as a Markov decision process in this paper. There are slight changes between poses in adjacent frames. When the pose in frame $t$ is given, we can roughly infer the corresponding pose in the next frame. Based on the above observation, the temporal correlation of the pose can be simplified into a Markov decision process \cite{DD-MCMC}, that is, the pose at the current moment is only related to the previous moment. Given the state at time $t$ (\emph{i.e.}, the human pose $m_{t}$), the state at time $t+1$ ($m_{t+1}$) can be denoted as follows. 
 \begin{equation}
    P\left(m_{t+1} | m_{t}, m_{t-1}, \cdots, m_{1}\right) = P\left(m_{t+1} | m_{t}\right)
    \label{eq:markvo}
 \end{equation}
 where $P(\cdot)$ is the state transition matrix. Eq. \ref{eq:markvo} shows that the state at time $t+1$ can be obtained based on the state at time $t$.
 
 For a series of sequential frames (\emph{i.e.}, $\mathcal{I}=\{I_{t}, I_{t+1},\cdots, I_{t+T}\}$) randomly sampled from a video, we take joints' heatmaps, $M=\left\{m_t, m_{t+1}, \cdots, m_{t+T}\right\}$, as states in $T$ continuous frames. We can obtain the observation $F(I_t)$ from frame $t$ via a function $F(\cdot)$. Based on the observation values of two adjacent frames, the target's latent motion $(\mathcal{O}_{t, t+1})$ can be modelled through a motion function $\phi(\cdot)$. Combining the motion information and the state at frame $t$ will preliminarily predict the state for the frame $t+1$. The predicted state is uncertain due to the lack of information about frame $t+1$. Therefore, the observation of frame $t+1$ ($F_{t+1}$) is used to balance the above uncertainty. And the accurate state of frame $t+1$ ($m_{t+1}$) can be formulated as follows.
 \begin{equation}
     \begin{aligned}
         & m_{t+1} = \psi{ \left( \zeta\left( \mathcal{O}_{t, t+1}, m_{t} \right) + F(I_{t+1}) \right)} \\
         & \mathcal{O}_{t, t+1} = \phi{\left(F(I_t), F(I_{t+1})\right)}
     \end{aligned}
     \label{eq:formation}
 \end{equation}
 where $\zeta(\cdot)$ represents the initial state prediction function. $\psi(\cdot)$ is the correction function to generate the correct state for frame $t+1$. $\phi(\cdot)$ is the motion function used to model the dynamics of the target in continuous frames. In practice, we take joints' heatmaps and appearance features as the states and observations. Function $\psi(\cdot)$, $\phi(\cdot)$, and $F(\cdot)$ are implemented by the convolutional blocks. Details are described in \emph{Subsection \ref{sec:section_B}}.
 
\subsection{KIMNet Model for Video-based Pose Estimation}
 \label{sec:section_B}
\begin{figure*}
   \centering
   \includegraphics[scale=0.3]{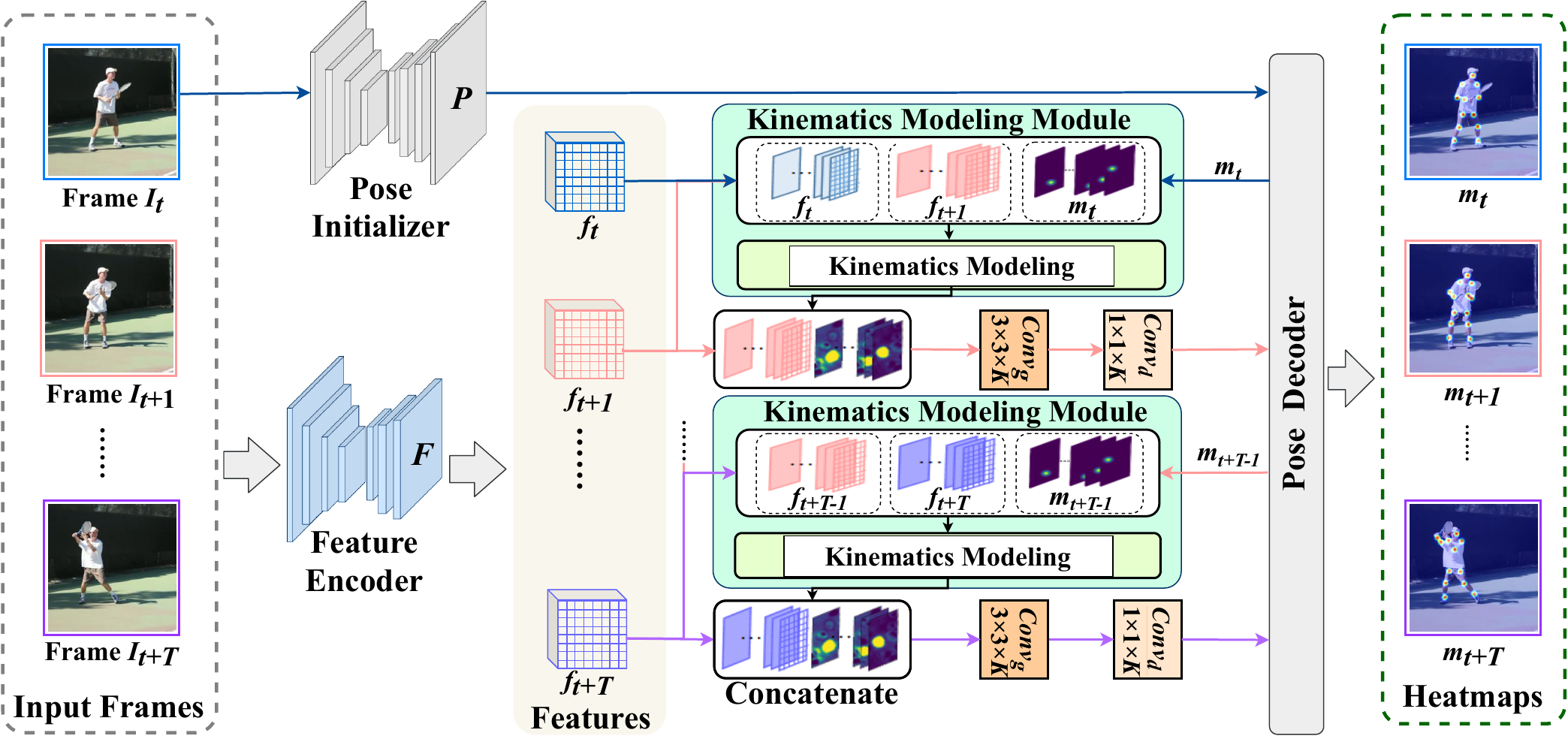}\\
   \caption{Overview of the proposed kinematics modeling network.}\label{fig:overview}
   %For the $1^{st}$ frame, the pose initializer $P$ and feature encoder $F$ are used to generate the initial heatmaps of joints and extract appearance features of poses. Then, the pose decoder (e.g. the JRE \cite{RPSTN}) is applied to produce the final heatmaps for joints. For subsequent frames, KMM takes pose features $f_t$, $f_{t+1}$, and heatmaps of joints $m_t$ in the $t$-th frame as inputs to predict the initial positions of joints $m_{t+1}^p$ for the $(t+1)$-th frame . Next, $f_{t+1}$ is used to eliminate uncertainties contained by $m_{t+1}^p$. Finally, the pose decoder is applied to generate the final pose.}\label{fig:overview}
 \end{figure*}

 We design a kinematics modeling network (KIMNet) to fit equation \ref{eq:formation}, as shown in Figure \ref{fig:overview}. As stated in eq. \ref{eq:formation}, there are four critical steps in KIMNet's modeling process, including acquisition of observations and initial state, motion modeling of joints, prediction of the state, and correction of the state. 

 \textbf{Acquisition of observations and initial state.} We use a feature encoder (\emph{i.e.}, $F$) based on the convolutional neural network to extract appearance features from each input frame as the observation (\emph{i.e.}, $f_t$). Moreover, we adopt a pretrained pose initializer to estimate the initial pose from the first frame as the initial state. For the convenience of description, we take the frame $(t+1)$ as the current frame and the frame $t$ as the historical frame in this paper.
 
 \textbf{Motion modeling of joints.} Human motion cues are significant for estimating poses from videos. If the motion cues are modeled effectively, it is possible to infer joints' positions in frame $t+1$ by combining the motion cues with the historical positions of joints. To extract robust motion cues, we propose a kinematic modeling module (KMM) to explore each joint's dynamical information by modeling the temporal correlation between any two joints. Then each joint's motion cues are combined with its historical position in frame $t$ to predict its positions in frame $t+1$.

 As mentioned above, the motion information is modeled by modeling the temporal similarity between joint features $f_t$ and $f_{t+1}$. Joints are represented by a set of feature maps consisting of $h \times w$ pixels during modeling process. The motion of the whole feature map, that is, the motion of all pixels should be modeled to represent the motion of the joint. Considering that the dot-product can measure the correlation \cite{RPSTN}. Furthermore, the dot-product is essentially a weighted sum that integrates the information of all pixels in the feature map. Thus, it can reflect the motion of the whole joint's feature map to some extent.
 \begin{equation}
 {O_{t,t + 1}} = \phi \left( {{f_t},{f_{t + 1}}} \right), t = 1,2,...,T - 1
 \label{eq:KMM_Offset}
 \end{equation}
 where $\phi(\cdot)$ is the motion function that is implemented based on the dot-product to model the temporal dependency between joints across frames by calculating the temporal similarity between any them. $O_{t,t+1} \in \mathbb{R}^{K \times K}$ reflects the movement of the joint in frame $t+1$ relative to all joints in frame $t$.%The softmax operation is then used to turn $O_{t,t+1}$ to the temporally correlational attention. Details can be found in subsection $C$.
 
 \textbf{Prediction of the state.} $O_{t,t+1}$ contains the motion information about each joint, and $m_t$ includes the positional information of joints in the $t$-th frame. By fusing the positional information and motion cues of joints, the model can preliminarily predict the joints' positions in frame $t+1$ as follows.
 \begin{equation}
 m_{t + 1}^p = \zeta (O_{t,t+1},{m_t}), t = 1,2, \ldots ,T - 1
 \label{eq:KMM_Mp}
 \end{equation}
 where $m_{t + 1}^p \in \mathbb{R}^{h \times w \times K}$ is the initial state, \emph{i.e.}, predicted joint heatmaps in frame $t+1$. $\zeta \left(\cdot \right)$ is the prediction function based on the dot-product. Since each element in $O_{t,t+1}$ reflects the motion degree of $j$-th joint $J_{t+1}^j$ in frame $t+1$ related to $J_{t}^i$ in frame $t$, the model can predict the joints' initial position $m^p_{t+1}$ by conducting the weighted sum between motion information $O_{t,t+1}$ and position information $m_t$.
 
 \textbf{Correction of the state.} To obtain precise joint heatmaps, we apply a pose decoder to produce the final heatmaps. For the first frame, the pose decoder is directly applied to the initial pose predicted by the pose initializer to get the final pose. 
 
 For subsequent frames, there are some uncertainties in predicted states $m_{t + 1}^p$ due to the lack of information for frame $t+1$. Therefore, we introduce the observation from frame $t+1$ (\emph{i.e.}, joint features $f_{t+1}$) into the current predicted state to alleviate these uncertainties. Practically, we adopt a correction function $\psi(\cdot)$ consisting of a $3 \times 3$ and a $1 \times 1$ convolutions (\emph{i.e.}, $Con{v_g}$ and $Con{v_d}$) to fuse the pose features and predicted poses.
 \begin{equation}
 \begin{split}
 & f_{t + 1}^{coarse} = \sum\limits_{i{\rm{ = }}1}^K {w_{g}^i{\rm{*}}m_{t + 1}^p}  + \sum\limits_{j = K + 1}^{D} w_g^j* {f_{t + 1}} \\
 & f_{t + 1}^{fine} = \sum\limits_{i = 1}^K w_d^i *f_{t + 1}^{coarse}\\
 \end{split}
 \label{eq:core_equa}
 \end{equation}
 where $f_{t+1}^{coarse} \in \mathbb{R}^{h \times w \times D}$ and $f_{t+1}^{fine} \in \mathbb{R}^{h \times w \times K}$ represent the coarse- and fine-grained pose representations, respectively. $w^{i}_g \in \mathbb{R}^{3 \times 3}$ and $w^i_d \in \mathbb{R}^{1 \times 1}$ are convolutional filters in $Con{v_g}$ and $Con{v_d}$. And $*$ represents the convolutional operation. In practice, KIMNet takes the concatenation both of $m_{t + 1}^p$ and $f_{t + 1}$ as the input, and conducts $Conv_{g}$ to generate the coarse pose feature, which is equivalent to the sum of performing $Conv_{g}$ on $m_{t + 1}^p$ and then on $f_{t + 1}$. After $Con{v_g}$ and $Con{v_d}$, we obtain the coarse-to-fine pose features.

 Finally, the pose decoder is used to produce the final heatmap for each joint.
 \begin{equation}
 {m_{t + 1}} = {\rm{Decoder}} \left( f_{t + 1}^{fine} \right), t = 1,2,...,T - 1
 \label{eq:sub_decoder}
 \end{equation}
where Decoder($\cdot$) is implemented by the joint relation extractor \cite{RPSTN}.
 
\subsection{Kinematics Modeling Module}

\begin{figure}[htbp]
   \centering
   \includegraphics[scale=0.3]{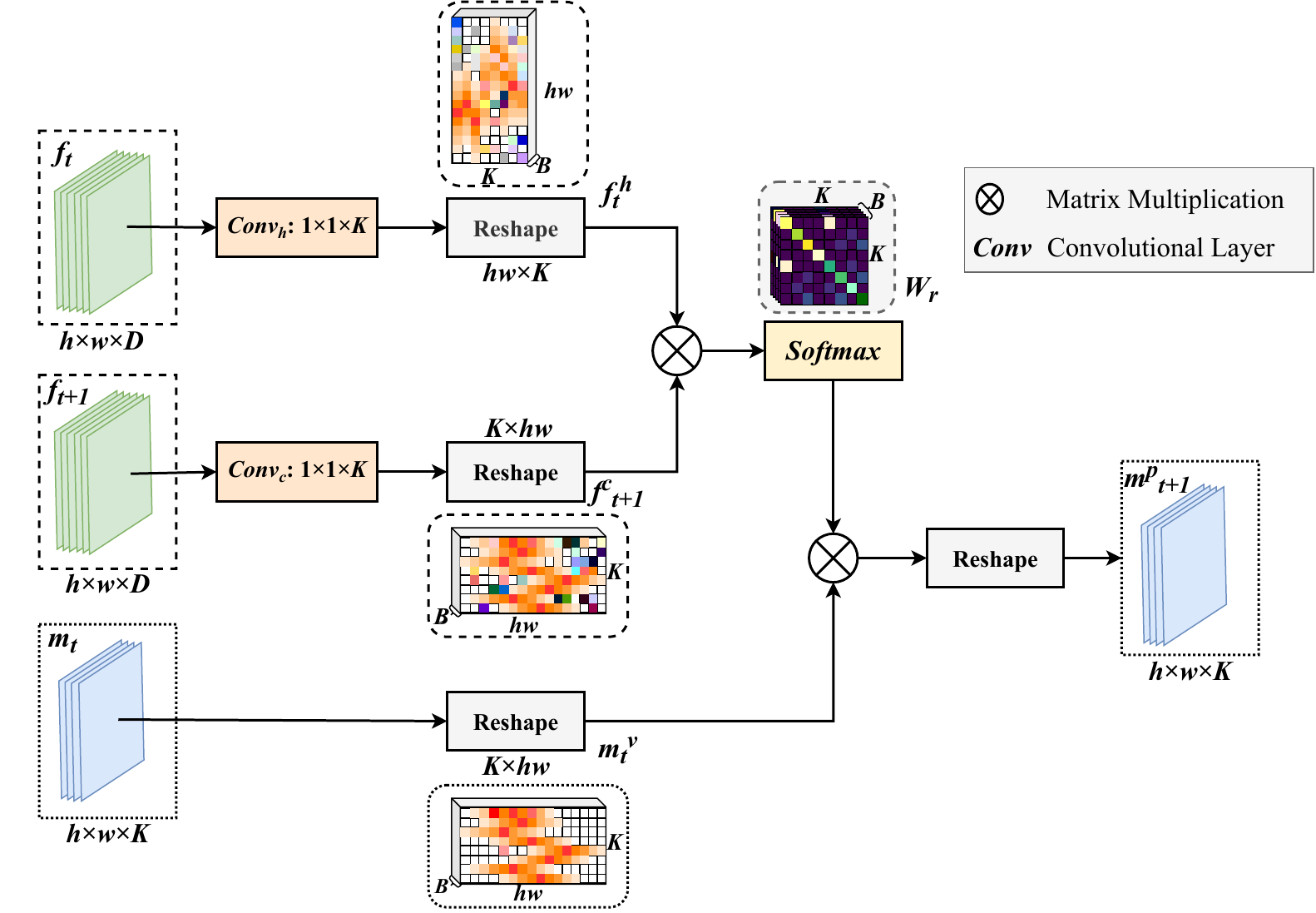}\\
   \caption{The structure of the proposed kinematics modeling module.}\label{fig:KMM}
 \end{figure}

 In order to model the temporal correlations between joints, we present a plug-and-play kinematics modeling module (KMM) based on the attention mechanism, as shown in Figure \ref{fig:KMM}. KMM fits the motion function $\phi(\cdot)$ and the initial state prediction function $\zeta(\cdot)$ in eq. \ref{eq:formation}, where the motion function $\phi(\cdot)$ contains two steps: robust pose representation extraction and temporal dependency modeling.

 \textbf{Robust pose representations extraction.} The information of the adjacent two frames is similar, resulting in the similarity between appearance feature $f_t$ and $f_{t+1}$. To explore robust motion information of poses, we adopt two independent $1 \times 1$ convolutions denoted as $Con{v_h}$ and $Con{v_c}$ to extract discriminative pose representations from $f_t$ and $f_{t+1}$. Then reshaping operation is used to convert the feature matrix into a feature vector for facilitating the temporal correlation modeling.
 \begin{equation}
    \begin{aligned}
        % f_t & = \left[f_t^1, f_t^2, \cdots, f_t^D \right] \in \mathbb{R}^{h \times w \times D}\\
        % f_{t+1} & = \left[f_{t+1}^1, f_{t+1}^2, \cdots, f_{t+1}^D \right] \in \mathbb{R}^{h \times w \times D}\\
        % W_h & = \left[w_h^1, w_h^2, \cdots, w_h^K \right] \in \mathbb{R}^{1 \times 1 \times K}\\
        % W_c & = \left[w_c^1, w_c^2, \cdots, w_c^K \right] \in \mathbb{R}^{1 \times 1 \times K}\\
        f_t^h & = {\mathop{\rm Reshape}\nolimits} \left(\sigma(W_h * f_t) \right) \in \mathbb{R}^{(h \times w) \times K}\\
        f_{t + 1}^c & = {\mathop{\rm Reshape}\nolimits}\left( \sigma(W_c * f_{t+1}) \right) \in \mathbb{R}^{K \times (h \times w)}
    \end{aligned}
 \label{eq:query_key}
 \end{equation}
 where $f_t \in \mathbb{R}^{h \times w \times D}$ and $f_{t+1} \in \mathbb{R}^{h \times w \times D}$ are joint features in frame $t$ and $t+1$. $W_h \in \mathbb{R}^{1 \times 1 \times K}$ and $W_c \in \mathbb{R}^{1 \times 1 \times K}$ are weights of $Con{v_h}$ and $Con{v_c}$. $\sigma(\cdot)$ and $*$ represents the nonlinear function and convolutional operation, respectively. Each column in $f_t^h$ and each row $f_{t+1}^c$ represent one joint.

 \textbf{Temporal dependency modeling.} Considering that joints are relevant to each other when the person is moving, we model the temporal correlation between any two joints by calculating their temporal similarity. Specifically, the dot-product can measure the similarity between two vectors to some extent. We apply the dot-product of pose features $f_t^h$ and $f_{t+1}^c$ to model the temporal correlation between any two joints across different frames. For the $i$-th and $j$-th joints represented by $f_t^{h_i} \in \mathbb{R}^{(h \times w) \times 1}$ and $f_{t+1}^{c_j} \in \mathbb{R}^{1 \times (h \times w)}$, the temporal correlation between them can be modeled as follows.
 \begin{equation}
 o_{t,t + 1}^{i,j} = f_{t + 1}^{{c_j}} \cdot f_t^{{h_i}}, t = 1,2, \ldots ,T - 1
 \label{eq:temp_corr}
 \end{equation}
 where ${o^{i,j}_{t,t + 1}}$ represents the temporal dependency between the joint $J_t^i$ and the joint $J_{t+1}^j$. Because the dot-product is essentially a weighted sum of joints' feature maps, it reflects the overall motion of the joint ${J^j_{t+1}}$ relative to $J_{t}^i$. Similarly, the temporal dependency between $J_{t+1}^j$ and other joints in frame $t$ can also be captured following Eq.\ref{eq:temp_corr}. This process can be realized by the matrix multiplication as follows.
 \begin{equation}
  O_{t, t+1} = \left( {\frac{{f_{t + 1}^c \otimes f_t^h}}{{\sqrt d }}} \right) \in \mathbb{R}^{K \times K},  t = 1,2, \ldots ,T - 1 
  \label{eq:joints_corr}
 \end{equation}
 where $\otimes$ denotes matrix multiplication used to model the temporal correlation between any two joints in different frames. $O_{t,t+1}=\{o^{1,1}_{t,t + 1}, o^{1,2}_{t,t + 1}, \cdots, \\o^{i,j}_{t,t + 1}, \cdots,o^{K,K}_{t,t + 1}\} \in \mathbb{R}^{K \times K}$ represents the temporal dependency between joints across frames $t$ and $t+1$. $d$ is the normalized factor that is equal to the dimension of features $f_t^h$ and $f_{t+1}^c$. 
 
 In order to enhance the motion cues of the joint $J_{t+1}^j$, the softmax operation, then, is used to model the global correlation \cite{GMFlow} that can be regarded as the dependency between the $J_{t+1}^j$ and all joints in the $t$-th frame.
 \begin{equation}
     w_r^{i,j} = \frac{1}{{\sum\limits_{l = 1}^K {o_{t,t + 1}^{l,j}} }} \cdot o_{t,t + 1}^{i,j},\begin{array}{*{20}{l}}
    {t = 1,2, \ldots ,T - 1}\\
    {i = 1,2, \ldots ,K}
    \end{array}
 \label{eq:correlation}
 \end{equation}
 where $w_r^{i,j}$ is the temporal attention weight between $J_{t+1}^j$ and $J_{t}^i$, reflecting the relative motion degree between two joints. $W_r=\{w_r^{1,1}, w_r^{1,2}, \cdots, w_r^{K,K}\} \in \mathbb{R}^{K \times K}$ highlights the attention of joints that are closely related to the current moving joint. Because $o_{t,t+1}^{i,j}$ includes the temporal dependency between two joints across different frames, the sum of ${o_{t,t + 1}^{l,j}}$ integrates temporal dependency of all joint pairs. Therefore, the Eq.\ref{eq:correlation} can reflect the motion information of $J_{t+1}^j$ relative to all joints in the $t$-th frame to some extent.

 \textbf{Prediction of joints' initial positions.} By fusing the known state (\emph{i.e.}, the historical joint's heatmap $m_t$) and the motion information, intuitively, the model is able to predict the initial state for frame $t+1$ (\emph{i.e.}, $m_{t+1}^p$) in advance. Since $W_r$ includes the motion information of joints, we apply $W_r$ to joint's position in frame $t$ to predict the $k$-th joint's position in frame $t+1$. The KMM computes the response at a position of $J_{t+1}^k$ as a weighted sum of the positions of all joints at frame $t$, which can be denoted as:
 \begin{equation}
 \begin{array}{l}
    \bar{m}_{t + 1}^{p} = W_r \cdot m_t^{v} \in \mathbb{R}^{K \times (h \times w)} 
 \end{array}
 \label{eq:infer}
 \end{equation}
 where $\bar{m}_{t + 1}^{p}$ is the initial heatmap vector of joints. In practice, we transfer the heatmap $m_t$ to ${m_t}^v \in \mathbb{R}^{K \times (h \times w)}$ for the convenient inference of $m_{t+1}^p$. The reshaping operation is used to convert $\bar{m}_{t + 1}^{p}$ to the initial heatmap $m_{t + 1}^{p}$ with the shape of $K \times h \times w$.

\subsection{Training Loss Function}
 In this paper, we minimize the $L_2$ norm between the output of the model and the heatmap of ground truth to optimize the proposed KIMNet. Specifically,
 \begin{equation}
 Loss = \frac{1}{T}\sum\limits_{t = 1}^T {\left\| {{M_t} - M{'_t}} \right\|_2^2}, t = 1,2, \ldots ,T
 \label{eq:loss}
 \end{equation}
 where $M_t=\{{m_t}^1, {m_t}^2, \ldots, {m_t}^K\}$ denotes the output of KIMNet. $M'_t=\{{m'_t}^1, {m'_t}^2, \ldots, {m'_t}^K\}$ is the joints' heatmap generated according to the ground truth. $T$ is the total number of training frames. $\left\|  \cdot  \right\|$ represents the $L_2$ norm.

\subsection{Comparisons with RPSTN}
\label{sec:difference_with_RPSTN}
 Although the structure of the proposed KIMNet is similar to \cite{RPSTN}, there are essential differences between them. \textbf{1. Motivations are different.} \cite{RPSTN} aims to explore the spatial affinity among joints by modeling the spatial correlations between joints within a frame. While the proposed KIMNet is designed to model the temporal dependency among joints across different frames. \textbf{2. Temporal modeling approaches are different.} \cite{RPSTN} uses several convolutional layers to transfer the historical pose knowledge and template matching to search for similar regions in the current frame, which ignores the temporal correlation between joints in different frames. In addition, as the number of convolution layers increases, the feature resolution decreases continuously, leading to the loss of historical pose information to a certain extent. In contrast to \cite{RPSTN}, the proposed KIMNet explicitly models the temporal correlation between joints across different frames. In this way, the model can locate the current joint by aggregating joints' information in the previous frame. Furthermore, the proposed KMM is implemented via attention mechanism, which preserves the original resolution of pose features while propagating historical features. Compared with RPSTN \cite{RPSTN}, KIMNet can avoid the loss of pose information due to the reduction of the resolution.

\section{Experiments}

We first introduce specific experimental settings. Then, we compare the proposed KIMNet against existing state-of-the-art video-based methods. We also provide ablation studies to confirm the effectiveness of the proposed KMM. Finally, we conduct comprehensive experiments in the occluded scene to show advantages of the proposed model. 

\subsection{Experimental Settings}

\subsubsection{Datasets}

 \textbf{Penn Action dataset} contains 2326 video clips. Each frame is annotated with 13 joints, including the head, shoulders, elbows, wrists, hips, knees, and ankles. Besides, the Penn Action dataset annotates the position of the person, \emph{i.e.}, the bounding box, and gives visibility for each joint. Keeping the consistency with \cite{DKD,RPSTN}, we train the model with 1258 video clips, and the others are used to evaluate the model.

 \textbf{Sub-JHMDB dataset} includes 316 videos. There are 15 body joints in each frame. Following \cite{RPSTN}, 3-fold-cross-validation is used to evaluate the model. According to existing methods, training sets include 227, 236, and 224 video clips, respectively, and testing sets correspondingly contain 89, 80, and 92 samples. To make a fair comparison with previous approaches, we take the average result on three testing sets as the final result.

\subsubsection{Implementation details}

 Following \cite{RPSTN}, SimpleBaseline \cite{SimpleBaseline} takes ResNet-101 as the backbone and is used as the pose initializer. SimpleBaseline \cite{SimpleBaseline} that takes ResNet-50 as the backbone is chosen as the feature encoder. Both pose initializer and feature encoder are pre-trained on the MPII \cite{MPII} dataset. Furthermore, to make a fair comparison with existing methods, keeping consistency with works \cite{DKD,RPSTN}, we also: randomly select 5 contiguous frames from each training video clip; adopt the same data augmentation settings as \cite{DKD,RPSTN} including random scaling, rotation, and flipping; use the Adam \cite{Adam} optimizer to train the model and the training epoch is set to 100 following \cite{RPSTN}. The learning rate is initialized to 0.005, and the batch size is set to 16 on the Penn Action dataset. Because the scale of the Sub-JHMDB dataset is smaller than that of Penn Action dataset, the learning rate and batch size are set to 0.001 and 8. All experiments are conducted on two NVIDIA GeForce RTX 3080Ti GPUs.

\subsubsection{Evaluation metric}

 The percentage of correct keypoints (PCK) is used as the evaluation metric. The PCK for each joint is calculated as follows.
 \begin{equation}
 PC{K_k} = \frac{{\sum\limits_{n = 1}^S {\sum\limits_{t = 1}^T {\eta \left( {\frac{{d_t^k}}{{{L_t}}} \le \alpha } \right)} } }}{S \cdot T}
 \label{eq:pck}
 \end{equation}
 where $PC{K_k}$ is the $k$-th joint's PCK. $S$ denotes the number of sequences; $T$ is the number of frames in each sequence. $\eta\left( \cdot \right)$ is the indicator function. When the condition in parentheses is true, $\eta\left( \cdot \right)$ is 1, otherwise it is 0. ${d_t}^{k}$ represents euclidean distance between the estimated value and the ground truth. $\alpha$ is the hyperparameter which is set to 0.2 following \cite{DKD,RPSTN} to control the range of errors. $L_t$ is the threshold of the error distance. According to \cite{DKD,RPSTN}, we set $L$ to the person size and the torso size, respectively.
 %: $L$ is set to the size of the person \cite{LSTMPose,DKD,RPSTN}, i.e., $L=max(H_{bbox}, W_{bbox})$, where $H_{bbox}$ and $W_{bbox}$ are the height and width of the bounding box of the person. 2) PCK normalized by torso size: $L$ is set as the size of the torso that is represented by the distance between the left shoulder and the right hip \cite{DKD,RPSTN}. 
 Based on the PCK of each joint, we also use the mean PCK (mPCK) to measure the overall performance of the model \cite{DKD,RPSTN}.
 \begin{equation}
 mPCK = \frac{{\sum\limits_{n = 1}^S {\sum\limits_{t = 1}^T {\sum\limits_{k = 1}^K {\eta \left( {\frac{{d_t^k}}{{{L_t}}} \le \alpha } \right)} } } }}{S \cdot T \cdot K}
 \end{equation}

\subsection{Comparison with State-of-the-arts}
\subsubsection{Results on the Penn Action dataset}
 \textbf{PCK normalized by person size.} We record the performance, parameters, and inference time of commonly used approaches, as shown in Table \ref{tab:pennaction}. The proposed KIMNet achieves the best performance on the PCK for each joint and mPCK The KIMNet outperforms the LSTMPM \cite{LSTMPose}, TCE \cite{TCE}, DKD \cite{DKD} and RPSTN \cite{RPSTN} by 2.0\%, 1.7\%, 1.9\%, and 1.0\% mPCK, respectively. These methods aggregate features along the temporal dimension but ignore the temporal correlations between joints. In contrast to these methods, the proposed KIMNet explicitly models the temporal correlations between joints, which is beneficial for locating joints. Furthermore, compared with the CNN-based methods, such as RPSTN \cite{RPSTN} and DKD \cite{DKD}, the attention mechanism makes KMM maintain the high resolution of feature maps, which can avoid the loss of spatial information caused by the reduction of the resolution. Besides, KIMNet abandons the cumbersome convolutional module that is used to model the temporal features of poses, so it spends less time estimating poses during inferencing.
 \begin{table*}[htbp]
  \footnotesize
  \setlength\tabcolsep{2pt}
  %\large
  \centering
  \caption{Comparisons with the state-of-the-art methods on the Penn Action dataset. Values in brackets indicate the gap between the corresponding approach and the proposed KIMNet.}
  \begin{tabular}{lcc|cccccccl}
   \hline
   Methods  & Params (M) & Time (ms) &  Head &  Sho. &  Elb. &  Wri. &  Hip &  Knee &  Ank. &  mPCK \\
   \hline
      &  &  & \multicolumn{8}{c}{Normalized by person size} \\
   Thin-slicing \cite{ThinSlicing} &
   - & - & 98.0 &  97.3 &  95.1 &  94.7 &  97.1 &  97.1 &  96.9 &  96.5 ($\downarrow$ 3.2) \\
   K-FPN \cite{KFPN} &
   - & - & 98.7 &  98.7 &  97.0 &  95.3 &  98.8 &  98.7 &  98.6 &  98.0 ($\downarrow$ 1.7) \\
   TCE \cite{TCE} &
   - & - & 99.3 &  98.5 &  97.6 &  97.2 &  98.6 &  98.1 &  97.4 &  98.0 ($\downarrow$ 1.7) \\
   UniPose \cite{UniPose} &
    - & - & - &  - &  - &  - &  - &  - &  - &  99.3 ($\downarrow$ 0.4) \\
   DCPose \cite{DCPose} &
   - & - &- & 98.6 & 96.2 & 96.0 & 98.7 & 98.8 & 98.7 & 97.9 ($\downarrow$ 1.8)   \\
   LSTMPM \cite{LSTMPose} &
   231.5 & 25 & 98.9 &  98.6 & 96.6 & 96.6 & 98.2 & 98.2 & 97.5 & 97.7 ($\downarrow$ 2.0) \\
   DKD \cite{DKD} &
   219.92 & 11 & 98.8 & 98.7 & 96.8 & 97.0 & 98.2 & 98.1 & 97.2 & 97.8 ($\downarrow$ 1.9) \\
   RPSTN \cite{RPSTN}  &
   222.17 & 12 & 99.0 & 98.7 & 98.8 & 98.5 & 98.8 & 98.7 & 98.8 & 98.7 ($\downarrow$ 1.0) \\
   \textbf{KIMNet}  &
   \textbf{214.70} & \textbf{10} & \textbf{99.4} & \textbf{99.6} & \textbf{99.1} & \textbf{99.0} & \textbf{99.8} & \textbf{99.5} & \textbf{99.4} & \textbf{99.7} \\

   \hline
      & &  \multicolumn{8}{c}{Normalized by torso size} \\

   LSTMPM \cite{LSTMPose} &
   231.5  & 25 & 96.0 & 93.6 & 92.4 & 91.1 & 88.3 & 94.2 & 93.5 & 92.6 ($\downarrow$ 4.3) \\
   DKD \cite{DKD} &
   219.92 & 11 & 96.6 & 93.7 & 92.9 & 91.2 & 88.8 & 94.3 & 93.7 & 92.9 ($\downarrow$ 4.0) \\
   RPSTN \cite{RPSTN} &
   222.17 & 12 & 98.2 & 96.9 & 95.2 & 93.2 & 96.6 & 95.7 & 95.0 & 95.7 ($\downarrow$ 1.2) \\
%   KIMNet (Res18) &
%   \textbf{126.26} & \textbf{8} & 97.2 & 97.1 & 95.0 & 93.9 & 96.1 & 96.2 & 94.8 & 95.6 ($\downarrow$ 1.3) \\
   \textbf{KIMNet} &
   \textbf{214.70} & \textbf{10} & \textbf{98.6} & \textbf{97.7} & \textbf{96.6} & \textbf{95.7} & \textbf{96.9} & \textbf{97.3} & \textbf{96.3} & \textbf{96.9} \\
   \hline
   \end{tabular}

   \label{tab:pennaction}
 \end{table*}

 \textbf{PCK normalized by torso size.} As shown in Table \ref{tab:pennaction}, our approach also ranks first among the existing methods with fewer parameters and inferencing time. And the overall performance of KIMNet is 1.2\% higher than that of RPSTN, which demonstrates the effectiveness of the KIMNet. In particular, we obtain encouraging improvements for those more challenging joints, such as \emph{wrists} and \emph{ankles}: with a PCK of 95.7\% $(\uparrow 1.8\%)$ for \emph{wrists} and a PCK of 96.3\% $(\uparrow 1.5\%)$ for \emph{ankles}, confirming the advantages of the KIMNet in locating the challenging joints. Experimental results prove that the temporal dependency between joints is beneficial for estimating poses from videos.

 \textbf{Compatibility of KMM.} We integrate KMM into various pose estimation frameworks, including LSTMPM \cite{LSTMPose}, DKD \cite{DKD}, and RPSTN \cite{RPSTN}, to evaluate its generality, as shown in Table \ref{tab:general}. Specifically, we replace the temporal modeling modules in raw backbones with the KMM. The performance of some joint points decreases slightly after integrating KMM, but the overall performance of the network is improved, which proves that the proposed KMM is compatible with the current popular video-based pose estimation frameworks.
 \begin{table}[htbp]
  \footnotesize
  \setlength\tabcolsep{2.5pt}
  %\large
  \centering
  \caption{Verification of KMM's generality on the Penn Action dataset. Evaluation metric is the PCK normalized by torso size.}
  \begin{tabular}{l|cccccccl}
   \hline
   Methods &  Head &  Sho. &  Elb. &  Wri. &  Hip &  Knee &  Ank. &  mPCK \\
   \hline
   LSTMPM \cite{LSTMPose} & 96.0 & 93.6 & \textbf{92.4} & 91.1 & 88.3 & 94.2 & \textbf{93.5} & 92.6 ($\downarrow$ 0.4) \\
   \textbf{LSTMPM-KMM} & \textbf{97.3} & \textbf{94.3} & 92.3 & \textbf{93.9} & \textbf{89.8} & \textbf{94.5} & 91.3 & \textbf{93.0} \\
   \hline
   DKD \cite{DKD} & 96.6 & 93.7 & \textbf{92.9} & 91.2 & 88.8 & 94.3 & \textbf{93.7} & 92.9 ($\downarrow$ 0.8) \\
   \textbf{DKD-KMM} & \textbf{97.7} & \textbf{96.4} & 91.5 & \textbf{96.2} & \textbf{89.7} & \textbf{95.4} & 91.4 & \textbf{93.7} \\
   \hline
   RPSTN \cite{RPSTN} & \textbf{98.2} & 96.9 & \textbf{95.2} & 93.2 & 96.6 & 95.7 & 95.0 & 95.7 ($\downarrow$ 0.2) \\
   \textbf{RPSTN-KMM} & 97.9 & \textbf{97.0} & 94.8 & \textbf{93.3} & 96.6 & \textbf{96.9} & \textbf{95.9} & \textbf{95.9} \\
   \hline
   \end{tabular}
   \label{tab:general}
   
 \end{table}

 \subsubsection{Results on the Sub-JHMDB dataset}
 We evaluate the KIMNet on the challenging Sub-JHMDB dataset to further confirm its effectiveness, as shown in Table \ref{tab:jhmdb}.

 \textbf{PCK normalized by person size.} We observe that previous approaches have achieved impressive results on the Sub-JHMDB dataset. KIMNet achieves 98.5\% mPCK that surpasses RPSTN \cite{RPSTN} by 1.1\% and outperforms SLT-Pose \cite{SLT-Pose} by 2.5\%. KIMNet explicitly models the temporal dependency between joints across different frames. In this way, KIMNet can locate the current joint via other joints correlated with the current joint. SLT-Pose \cite{SLT-Pose} models the temporal correlation between poses in different frames, but it only considers the pose-level correlations while ignoring the temporal correlation between joints.

 \textbf{PCK normalized by torso size.} Similar to the results on the Penn Action dataset, KIMNet also achieves the best performance. The advantages of our model over RPSTN \cite{RPSTN} are more significant, achieving over 2.3\% mPCK improvement. Especially for the joints with high flexibility, such as \emph{elbows}, \emph{wrists}, \emph{knees}, and \emph{ankles}, our model achieves 5.0\%, 5.3\%, 2.9\%, and 2.4\% improvements over the RPSTN \cite{RPSTN}. The reason is that KIMNet can explicitly model the temporal dependency between joints across frames while maintaining the original resolution of the feature maps, which makes KIMNet locate one joint according to rich auxiliary information of other joints. However, KIMNet shows inferior performance for the head and hip joints against RPSTN \cite{RPSTN}. The possible reason is that a person's scale in the Sub-JHMDB dataset is small. Furthermore, the head and hip move smoothly relative to other joints. It is possible to locate these two joints through their spatial and temporal features. RPSTN \cite{RPSTN} adopts a CNN-based structure to model the dynamics of poses. Compared to the KMM containing two 1x1 convolutions, the CNN-based module can extract more discriminative features about the head and hip.
 \begin{table}[htbp]
  \footnotesize
  \setlength\tabcolsep{2.5pt}
  %\large
  \centering
  \caption{Comparisons with the state-of-the-art methods on the Sub-JHMDB dataset. Values in brackets indicate the gap between the corresponding approach and the proposed KIMNet.}
  \begin{tabular}{l|cccccccl}
   \hline
   Methods &  Head &  Sho. &  Elb. &  Wri. &  Hip &  Knee &  Ank. &  mPCK \\
   \hline
      & \multicolumn{8}{c}{Normalized by person size} \\

   % ACPS \cite{PoseAction} &
   % 90.3 &  76.9 &  59.3 &  55.0 &  85.9 &  76.4 &  73.0 &  73.8 ($\downarrow$ 24.7) \\
   Thin-slicing \cite{ThinSlicing} &
   97.1 &  95.7 &  87.5 &  81.6 &  98.0 &  92.7 &  89.8 &  92.1 ($\downarrow$ 6.4) \\
   K-FPN \cite{KFPN} &
   95.1 &  96.4 &  95.3 &  91.3 &  96.3 &  95.6 &  92.6 &  94.7 ($\downarrow$ 3.8) \\
   TCE \cite{TCE} &
   99.3 &  98.9 &  96.5 &  92.5 &  98.9 &  97.0 &  93.7 &  96.5 ($\downarrow$ 2.0) \\
   LSTMPM \cite{LSTMPose} &
   98.2 &  96.5 & 89.6  &  86.0 &  98.7 &  95.6 &  90.9 &  93.6 ($\downarrow$ 4.9) \\
   DKD \cite{DKD} &
   98.3 &  96.6 & 90.4  &  87.1 &  99.1 &  96.0 &  92.9 &  94.0 ($\downarrow$ 4.5) \\
   DCPose \cite{DCPose} &
   - & 97.9 & 93.2 & 92.1 & 98.4 & 97.8 & 95.9 & 95.8 ($\downarrow$ 2.7) \\
   
   FAMI-Pose \cite{TFAM} & \textbf{99.3} & 98.6 & 94.5 & 91.7 & \textbf{99.2} & 91.8 & 95.4 & 96.0 ($\downarrow$ 2.5)\\

   SLT-Pose \cite{SLT-Pose} & \textbf{99.3} & 98.6 & 94.3 & 91.4 & \textbf{99.2} & 91.9 & 95.3 & 96.0 ($\downarrow$ 2.5)\\
   
   RPSTN \cite{RPSTN}  &
   98.9 & \textbf{99.1} & \textbf{99.0}  &  97.9 &  97.8 &  97.8 &  \textbf{97.3} &  97.4 ($\downarrow$ 1.1) \\
   \textbf{KIMNet (Ours)}  &
   98.9 & \textbf{99.1} & 98.7 & \textbf{98.5} & 98.6 & \textbf{98.3} & 97.2 & \textbf{98.5} \\
    
   \hline
      & \multicolumn{8}{c}{Normalized by torso size} \\

   LSTMPM \cite{LSTMPose} &
   92.7 & 75.6 & 66.8 & 64.8 & 78.0 & 73.1 & 73.3 & 73.6 ($\downarrow$ 14.5) \\   
   DKD \cite{DKD} &
   \textbf{94.4} & 78.9 & 69.8 & 67.6 & 81.8 & 79.0 & 78.8 & 77.4 ($\downarrow$ 10.7) \\
   RPSTN \cite{RPSTN} &
   91.0 & 87.1 & 82.1 & 80.5 & \textbf{88.8} & 85.9 & 83.8 & 85.8 ($\downarrow$ 2.3) \\
   \textbf{KIMNet (Ours)} &
   90.1 & \textbf{88.8} & \textbf{87.1} & \textbf{85.8} & 87.9 & \textbf{88.8} & \textbf{86.2} & \textbf{88.1} \\
   \hline
   \end{tabular}

   \label{tab:jhmdb}
 \end{table}

\subsection{Ablation Studies}

 We first evaluate the effectiveness of the proposed method from the temporal modeling method. Then we change the internal structure and the resolution of KMM to verify the validity of KMM. Finally, we validate the influence of the feature encoder.

\subsubsection{Ablation studies about temporal modeling method}

 To demonstrate the effectiveness of the KMM, we compare the proposed KMM with previous temporal modeling modules, as shown in Table \ref{tab:ablation2_kmm}. We first remove the temporal modeling module, \emph{i.e.}, the Baseline. Obviously, compared with our KIMNet, the model's performance degrades significantly after removing the temporal modeling, which proves the necessity of temporal modeling for video-based human pose estimation.
 
 Second, to evaluate the predictive performance of KMM, we remove the features for frame $t+1$ (\emph{i.e.}, $F\left(I_{t+1}\right)$ in eq. \ref{eq:formation}, and denote it as ``Predicted" in Table \ref{tab:ablation2_kmm}). Experimental results show that KMM provides acceptable prediction results. After integrating the observations of the current frame, the model obtains accurate positions of joints, demonstrating the necessity of $F\left(I_{t+1}\right)$ for refining joints' positions.
 
 Third, we adopt the DKD proposed in \cite{DKD} and JRPSP introduced by \cite{RPSTN} to model the temporal information. Specifically, we successively replace the KMM in KIMNet with DKD and JRPSP. Compared with DKD and JRPSP, KMM brings a 1.1\% and 0.9\% improvement of overall performance. Especially for the flexible \emph{wrists} and \emph{ankles}, the improvements are obvious. These joints generally move violently and are easily prone to motion blur. It is helpful to locate these joints by fusing other joints' information. 
 \begin{table}[htbp]
   \footnotesize
   \setlength\tabcolsep{3pt}
   %\large
   
   \centering
   \caption{Ablation studies about the temporal modeling module. Evaluation metric is the PCK normalized by torso size. Values in brackets indicate the gap between the corresponding approach and our method.}
   \begin{tabular}{l|cccccccl}
   \hline
   Method &  Head &  Sho. &  Elb. &  Wri. &  Hip &  Knee &  Ank. &  mPCK \\
   \hline
    Baseline & 97.5 & 96.7 & 93.1 & 91.2 & 96.0 & 95.1 & 93.7 & 94.5 ($\downarrow$ 2.4) \\
    Predicted & 93.0 & 94.5 & 92.3 & 85.1 & 95.2 & 90.0 & 89.3 & 91.2 ($\downarrow$ 5.7) \\
    DKD \cite{DKD} & 97.2 & 97.0 & 94.6 & 93.6 & 96.3 & 96.8 & 95.4 & 95.8 ($\downarrow$ 1.1) \\
    JRPSP \cite{RPSTN} & 97.5 & 97.2 & 96.2 & 92.5 & 97.1 & 96.9 & 95.5 & 96.0 ($\downarrow$ 0.9) \\
   \hline
   \textbf{KIMNet} & \textbf{98.6} & \textbf{97.7} & \textbf{96.6} & \textbf{95.7} & \textbf{96.9} & \textbf{97.3} & \textbf{96.3} & \textbf{96.9} \\
   \hline
   \end{tabular}
   \label{tab:ablation2_kmm}
   
 \end{table}

% \subsubsection{Ablation studies about the correction function}}
%  We remove the correction function to evaluate the predicted poses by the KMM. As shown in Table \ref{tab:correction}, "Predicted" means that we remove the observations of the current frame and calculate the evaluation metric between the predicted heatmaps and ground truth. Experimental results show that it is necessary to introduce the observation of the current frame to reduce the uncertainty of the prediction, so as to get the accurate poses from the current frame. Especially for those complex actions, KMM is difficult to provide accurately predicted heatmaps for them, as shown in Figure \ref{fig:motion_result}.}
% \begin{table}[htbp]
%   \footnotesize
%   \setlength\tabcolsep{3pt}
%   %\large
%   
%   \centering
%   \caption{Ablation studies about the correction function. Evaluation metric is the PCK normalized by torso size. Values in brackets indicate the gap between the corresponding approach and our method.}
%   \begin{tabular}{l|cccccccl}
%   \hline
%   Method &  Head &  Sho. &  Elb. &  Wri. &  Hip &  Knee &  Ank. &  mPCK \\
%   \hline
%     Predicted & 93.0 & 94.5 & 92.3 & 85.1 & 95.2 & 90.0 & 89.3 & 91.2 ($\downarrow$ 5.7) \\
%   \textbf{KIMNet (Ours)} & \textbf{98.6} & \textbf{97.7} & \textbf{96.6} & \textbf{95.7} & \textbf{96.9} & \textbf{97.3} & \textbf{96.3} & \textbf{96.9} \\
%   \hline
%   \end{tabular}
%   \label{tab:correction}
%   }
%  \end{table}

\subsubsection{Ablation studies about KMM's structure}
 
 We have carried out a comprehensive analysis about the structure of KMM. Experimental results are recorded in Table \ref{tab:ablation3_kmm}. 
 
 \textbf{Influence of shared $Con{v_h}$ and $Con{v_c}$.} We adopt two $1 \times 1$ convolutions whose weights are shared to extract pose representations from two adjacent frames, \emph{i.e.}, KIMNet-Shared-hc in Table \ref{tab:ablation3_kmm}. As can be seen that the performance of KIMNet-Shared-hc decreases by 1.1\% mPCK over that of KIMNet. Two independent convolutions can extract discriminative pose features from two contiguous frames, which is beneficial for capturing the robust motion information of joints. 
 
 \textbf{Influence of historical heatmaps.} We introduce the residual, \emph{i.e.}, heatmaps in frame $t$, into KMM for evaluating the influence of historical position for predicting joints' positions, \emph{i.e.}, KIMNet-Residual. The performance decreases obviously after introducing the historical position of joints. We think that $m_{t+1}^p$ has included the positional information about joints in frame $(t+1)$. The model pays much attention to the historical position after introducing historical positions. 
 
 \textbf{Temporal correlations between heatmaps.} We adopt heatmaps in two frames to model the temporal correlation between joints, \emph{i.e.}, KIMNet-Heatmap. The performance of KIMNet-Heatmap also decreases a lot. Compared with joints' heatmaps which include positional information, pose features $f$ contain rich semantic information about joints, including appearance features and positional information, which is helpful for learning the robust motion cues.
 \begin{table}[htbp]
   \footnotesize
   \setlength\tabcolsep{2.5pt}
   %\large
   \centering
   \caption{Ablation studies about the rationality of KMM's design. Evaluation metric is the PCK normalized by torso size. Values in brackets indicate the gap between the corresponding approach and our method.}
   \begin{tabular}{l|cccccccl}
   \hline
   Method &  Head &  Sho. &  Elb. &  Wri. &  Hip &  Knee &  Ank. &  mPCK \\
   \hline
   KIMNet-Share-hc &
   97.4 & 97.1 & 95.0 & 93.1 & 96.8 & 96.7 & 95.1 & 95.8 ($\downarrow$ 1.1) \\
   KIMNet-Residual &
   97.4 & 95.7 & 88.9 & 81.1 & 94.8 & 91.8 & 85.9 & 90.3 ($\downarrow$ 6.6) \\
   KIMNet-Heatmap &
   96.2 & 94.9 & 86.3 & 82.5 & 95.6 & 92.8 & 87.6 & 90.4 ($\downarrow$ 6.5) \\
   \hline
   \textbf{KIMNet} &
   \textbf{98.6} & \textbf{97.7} & \textbf{96.6} & \textbf{95.7} & \textbf{96.9} & \textbf{97.3} & \textbf{96.3} & \textbf{96.9} \\
   \hline
   \end{tabular}

   \label{tab:ablation3_kmm}
 \end{table}

\subsubsection{Ablation studies about resolution of propagating features}
The attention mechanism allows KMM to maintain the original resolution of pose features during modeling the temporal dependency between joints. We gradually reduce the resolution of KMM's input to verify the importance of retaining the original resolution during propagating pose knowledge. As shown in Table \ref{tab:resolution}, the performance of the KIMNet degrades with the reduction of the resolution, demonstrating that the significance of the high resolution for pose features.% Moreover, the locating accuracy degrades significantly for those joints that move relatively vigorously, such as the \emph{wrist} and \emph{ankle}. The possible reason is that these joints easily suffer from motion blur and occlusion since they move vigorously. Rich pose features are helpful to assist in locating these joints.
\begin{table}[htbp]
    \footnotesize
    \setlength\tabcolsep{3.6pt}
    \centering
    
    \caption{Ablation studies about the resolution of KMM's input. Evaluation metric is the PCK normalized by torso size. ``KIMNet-\emph{S}'' represents the size of KMM's input is $S \times S$. Values in brackets indicate the gap between the corresponding approach and our method. }
   \begin{tabular}{l|cccccccl}
   \hline
   Method &  Head &  Sho. &  Elb. &  Wri. &  Hip &  Knee &  Ank. &  mPCK \\
   \hline
   KIMNet-8 &
   96.0 & 94.8 & 88.2 & 81.1 & 95.9 & 93.1 & 88.9 & 90.8 ($\downarrow$ 6.1) \\
   KIMNet-16 &
   96.6 & 95.3 & 90.9 & 85.8 & 96.6 & 94.6 & 90.5 & 92.4 ($\downarrow$ 4.5) \\
   KIMNet-32 &
   97.3 & 95.8 & 91.0 & 87.0 & 97.2 & 96.1 & 91.7 & 93.5 ($\downarrow$ 3.4) \\
   \hline
   \textbf{KIMNet-64} &
   \textbf{98.6} & \textbf{97.7} & \textbf{96.6} & \textbf{95.7} & \textbf{96.9} & \textbf{97.3} & \textbf{96.3} & \textbf{96.9} \\
   \hline
   \end{tabular}
   \label{tab:resolution}
\end{table}

\subsubsection{Ablation studies about feature encoder}
  We adopt various models, including SimpleBaseline \cite{SimpleBaseline} with different structures (ResNet-18, ResNet-34, and ResNet-50), Hourglass \cite{Hourglass}, and HRNet \cite{HRNet} as the feature encoder to evaluate the influence of different backbones. Feature encoders are pre-trained on the MPII dataset \cite{MPII}. Experimental results are listed in Table \ref{tab:ablation1_PF}. For different feature encoders, the performance of KIMNet is improved with the enhancement of the feature extraction ability of the feature encoder. Compared to the SimpleBaseline with ResNet-50 as the backbone, \emph{i.e.}, ${\rm KIMNet}^{\dag}$, Hourglass \cite{Hourglass} (KIMNet(HG)) and HRNet (KIMNet(HRNet)) bring limited improvement. Considering the performance and computational cost, the SimpleBaseline with ResNet-50 as the backbone is chosen as the feature encoder in this paper.

 \begin{table}[htbp]
   \footnotesize
   \setlength\tabcolsep{2.5pt}
   %\large
   \centering
   \caption{Ablation studies about the feature encoder. Evaluation metric is the PCK normalized by torso size. ${\rm KIMNet}^{\dag}$ represents the structure used in this paper. Values in brackets indicate the gap between the corresponding approach and the proposed ${\rm KIMNet}^{\dag}$.}
   \begin{tabular}{l|cccccccl}
   \hline
   Methods &  Head &  Sho. &  Elb. &  Wri. &  Hip &  Knee &  Ank. &  mPCK \\
   \hline
  % KIMNet-w/o-P &
%   97.9 & 97.1 & 95.7 & 95.1 & 96.6 & 96.9 & 95.5 & 96.3 ($\downarrow$ 0.6) \\
   KIMNet (Res18) &
   97.2 & 97.1 & 95.0 & 93.9 & 96.1 & 96.2 & 94.8 & 95.6 ($\downarrow$ 1.3) \\
   KIMNet (Res34) &
   97.6 & 97.3 & 95.4 & 93.3 & 96.2 & 96.4 & 95.0 & 95.8 ($\downarrow$ 1.1) \\
   KIMNet (HG) &
   98.4 & 97.7 & 96.8 & 95.6 & 97.2 & \textbf{97.6} & \textbf{97.1} & 97.1 ($\uparrow$ 0.2) \\
   KIMNet (HRNet) &
   98.0 & \textbf{97.8} & \textbf{97.2} & \textbf{95.7} & \textbf{97.3} & 97.5 & \textbf{97.1} & \textbf{97.2} ($\uparrow$ 0.3) \\
   ${\rm KIMNet}^{\dag}$ (Res50) &
   \textbf{98.6} & 97.7 & 96.6 & \textbf{95.7} & 96.9 & 97.3 & 96.3 & 96.9 \\
   \hline
   \end{tabular}

   \label{tab:ablation1_PF}
 \end{table}

\subsection{Qualitative Analysis about Features Learned by KMM}
\subsubsection{Visualization of temporal correlation}
\begin{figure}[htbp]
    \centering
    \includegraphics[scale=0.15]{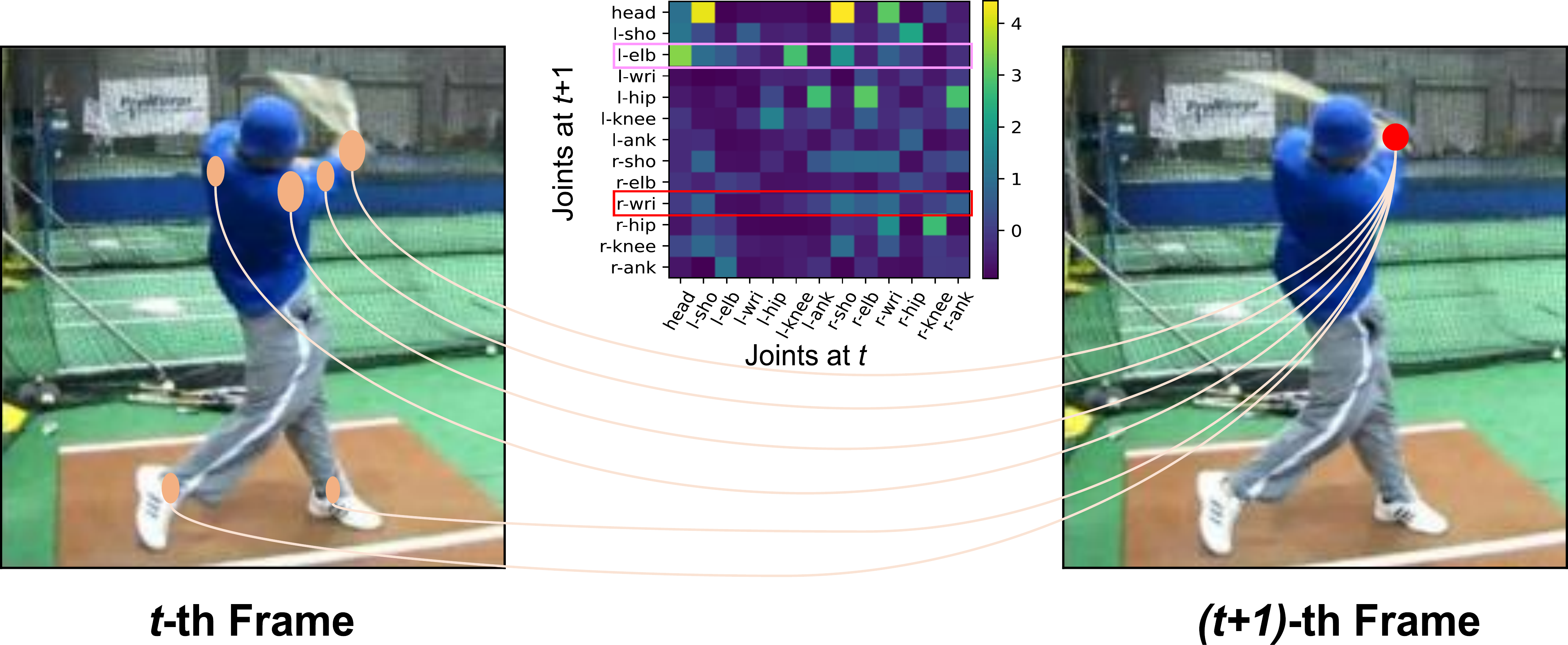}
    \caption{Visualization of the temporal correlation between joints. The size of joints at frame $t$ represents the degree of temporal dependency between joints.}
    \label{fig:correlation}
 \end{figure}

 To intuitively observe the temporal dependency between joints at different times, we randomly visualize two continuous frames and the corresponding temporal correlation matrix, as shown in Figure\ref{fig:correlation}. The size of joints at frame $t$ represents the degree of temporal dependency. The \emph{right wrist} has a high temporal dependency with the \emph{right wrist}, \emph{right elbow}, and \emph{shoulders} in frame $t$. At the same time, the \emph{left} and \emph{right ankles} work as the support foot and the force generating foot, respectively, and they cooperate with the \emph{right wrist} to complete the action while ensuring the coordination of the action. Therefore, there is a temporal dependency between the \emph{right wrist} and two \emph{ankles} in frame $t$. These joints provide extra temporal information to help locate the \emph{right wrist} from frame $t+1$. Similarly for the occluded \emph{left elbow} at frame $t+1$, the auxiliary temporal information provided by the \emph{head}, \emph{shoulders}, \emph{wrists}, and \emph{left knee} at frame $t$ helps the model to locate \emph{left elbow} from frame $t+1$. Therefore, with the help of the temporal dependency between joints, the proposed KIMNet uses more extra temporal information provided by other joints to locate the current joint rather than only depending on the current joint’s information to locate it.

\subsubsection{Visualization of joint heatmaps}

\begin{figure*}[htbp]
    \centering
    \subfloat[Visualization for simple poses.]{
        \includegraphics[width=0.47\linewidth]{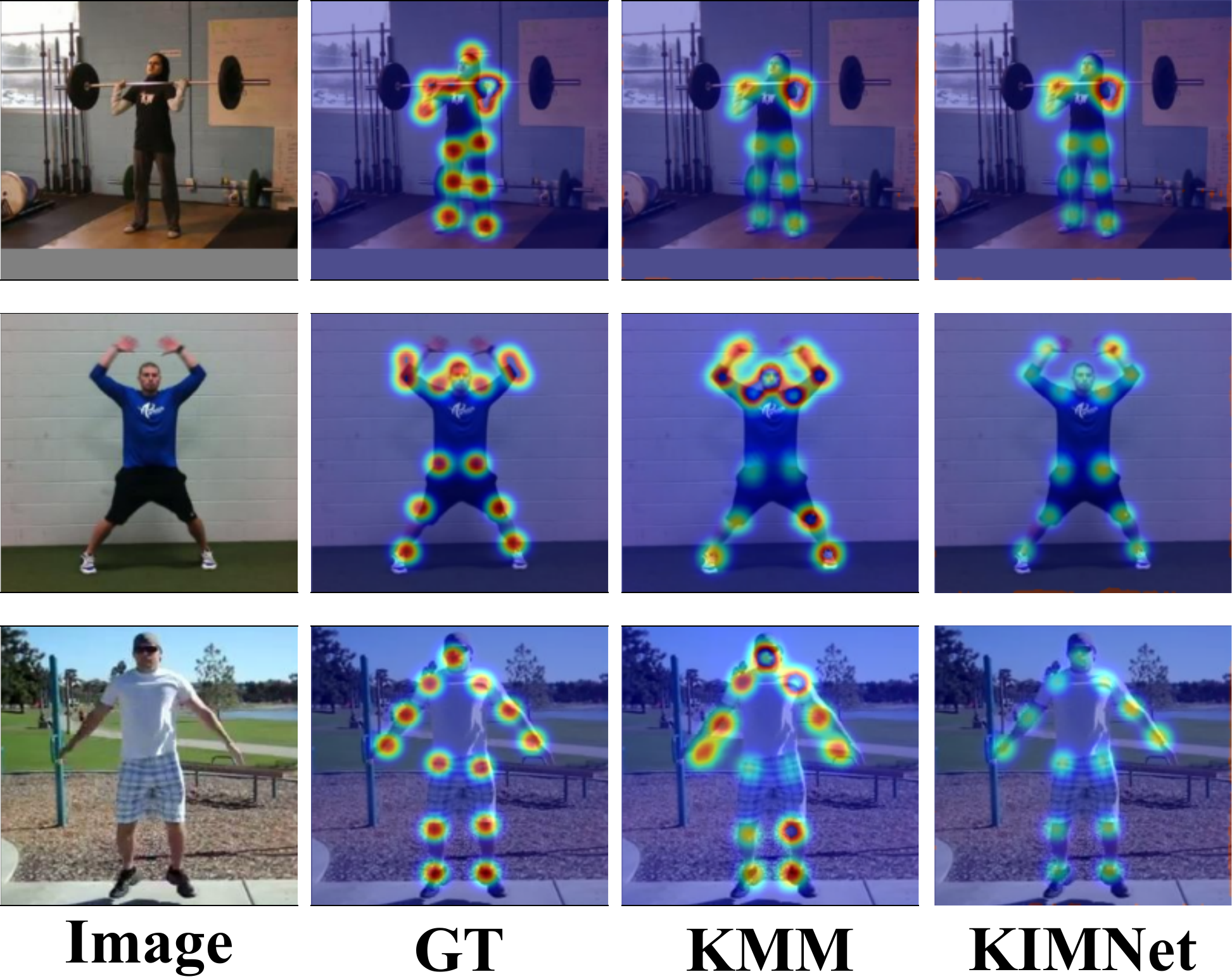}
        \label{fig:simple_pose}
    }
    \subfloat[Visualization for complex poses.]{
        \centering
        \includegraphics[width=0.475\linewidth]{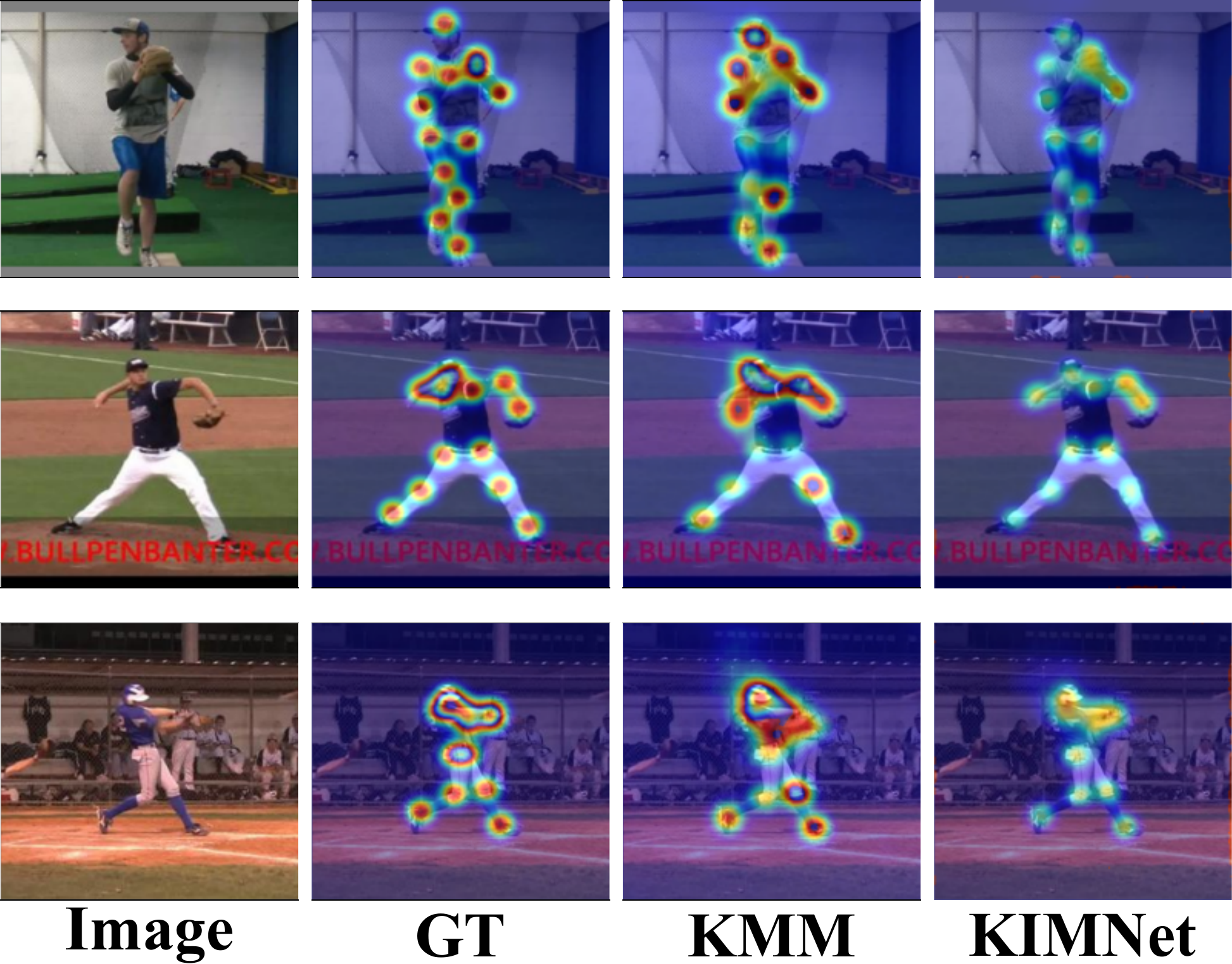}
        \label{fig:complex_pose}
    }%
   \caption{Outputs of the proposed KMM and KIMNet. The visualization results from left to right are the original input frames, ground truth, outputs of the KMM, and outputs of the KIMNet.}
   \label{fig:occlud_vis}
\end{figure*}

 To further validate the effectiveness of KIMNet, we visualize the predicted maps of KMM and heatmaps of KIMNet for different poses with various complexity, and display heatmaps of ground truth ($2^{nd}$ column), outputs of KMM (the $3^{rd}$ column), and KIMNet (the $4^{th}$ column) in Figure \ref{fig:occlud_vis}. Specifically, we randomly select one frame from the $2^{nd}$ frame to the $T$-th frame and visualize heatmaps of joints. 
 
 We observe that KMM can provide accurate initial positions of joints for those simple actions, as shown in Figure \ref{fig:simple_pose}, actions \emph{weight lifting} and \emph{jumping jack}. Joints move homogeneously in these simple actions. Thus, it is easy for KMM to model the temporal dependency between joints, which makes KMM learn obvious motion cues for these joints. Moreover, KMM can guide the model to pay much attention to the posing area for those complex actions, as shown in Figure \ref{fig:complex_pose}. For the action \emph{play baseball}, joints move violently, especially the joints of the upper limbs. Therefore, KMM pays much attention to the upper limb, indicating that joints of the upper limbs have similar movements. Based on the regions activated by the KMM, KIMNet can locate joints accurately.

\subsection{Verification of Performance in Occlusion Scenes}
 To explore the potential of the proposed KIMNet in occluded scenes, we evaluate the KIMNet in two cases of the occlusion, including the manually occluded scenario and the naturally occluded scenario.
\begin{figure*}[htbp]
    \subfloat{
        \includegraphics[width=0.47\linewidth]{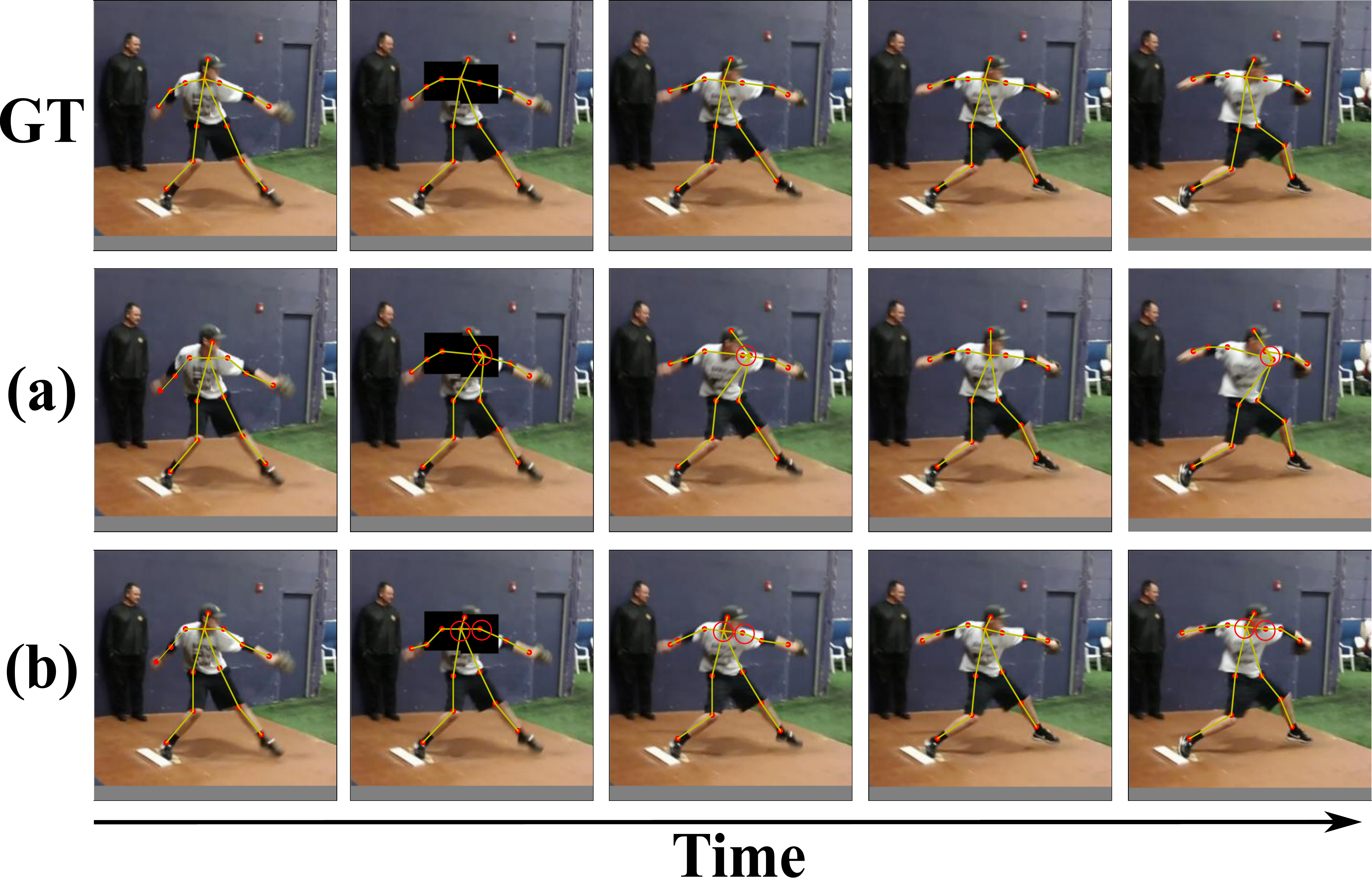}
        \label{fig:temporal_occlude_subfig_1}
    }
    \subfloat{
        \includegraphics[width=0.475\linewidth]{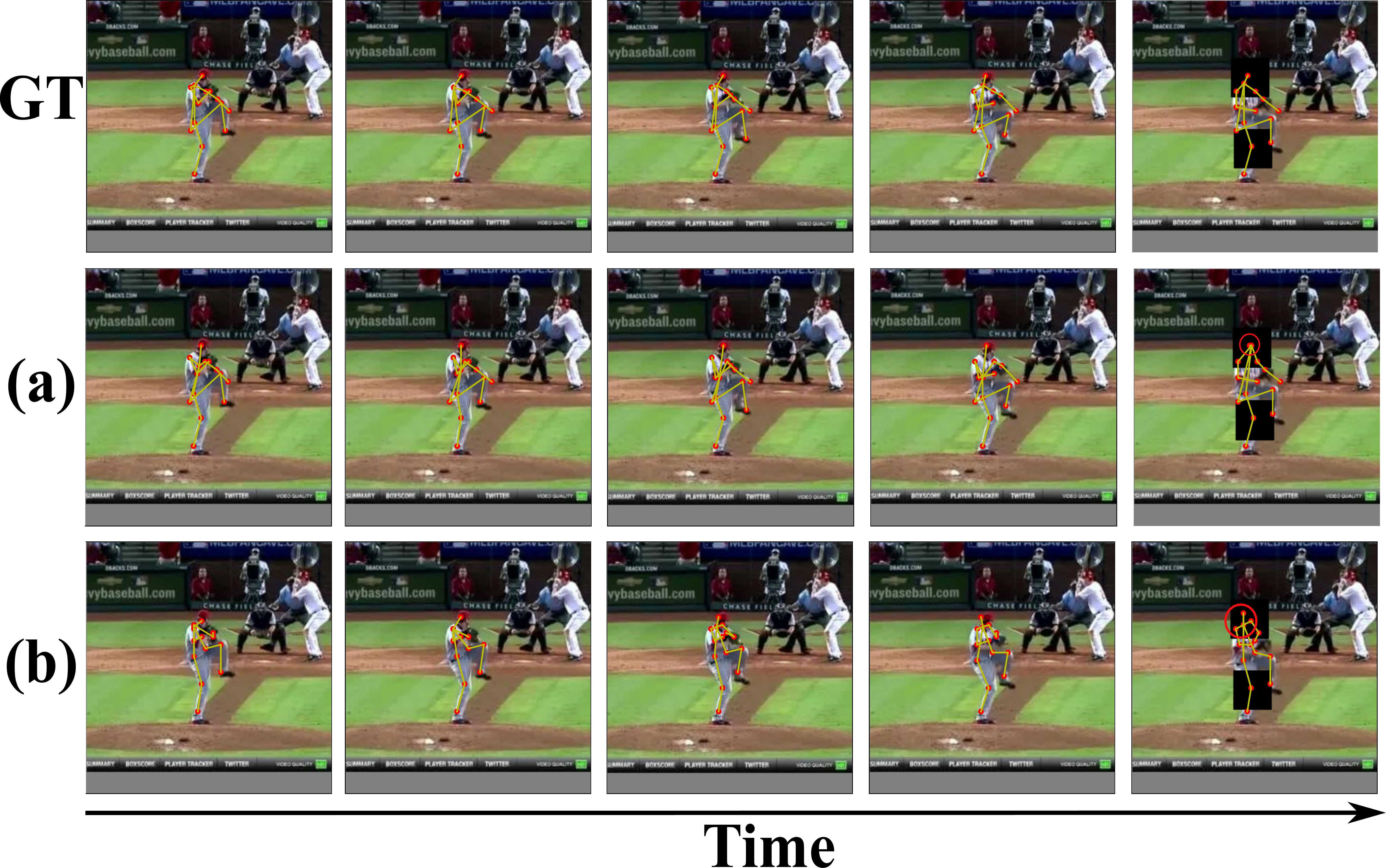}
        \label{fig:temporal_occlude_subfig_2}
    }%
   \caption{Visualization of the temporal occlusion. (a) is the output of RPSTN \cite{RPSTN}. (b) is the output of KIMNet. We use red circles to highlight the joints that are localized more accurate by KIMNet.}
   \label{fig:temporal_occlude}
\end{figure*} 
 
 \subsubsection{The Manually Occluded Scenario}
 We randomly generate several masks with a size of $40 \times 40$ from time and space to occlude joints for evaluating KIMNet's performance against the temporal and spatial occlusion, respectively.
 
 \textbf{Verification of temporal occlusion.} We randomly select a frame from the input sequence and generate two masks to occlude joints. Experimental results of the proposed KIMNet and RPSTN are shown in Figure\ref{fig:temporal_occlude}. Because our proposed KIMNet explicitly models the temporal dependency between joints across different frames, it can use all joints in the previous frame to help locate joints in the current frame.
 
\begin{figure*}[htbp]
    \subfloat{
        \includegraphics[width=0.47\linewidth]{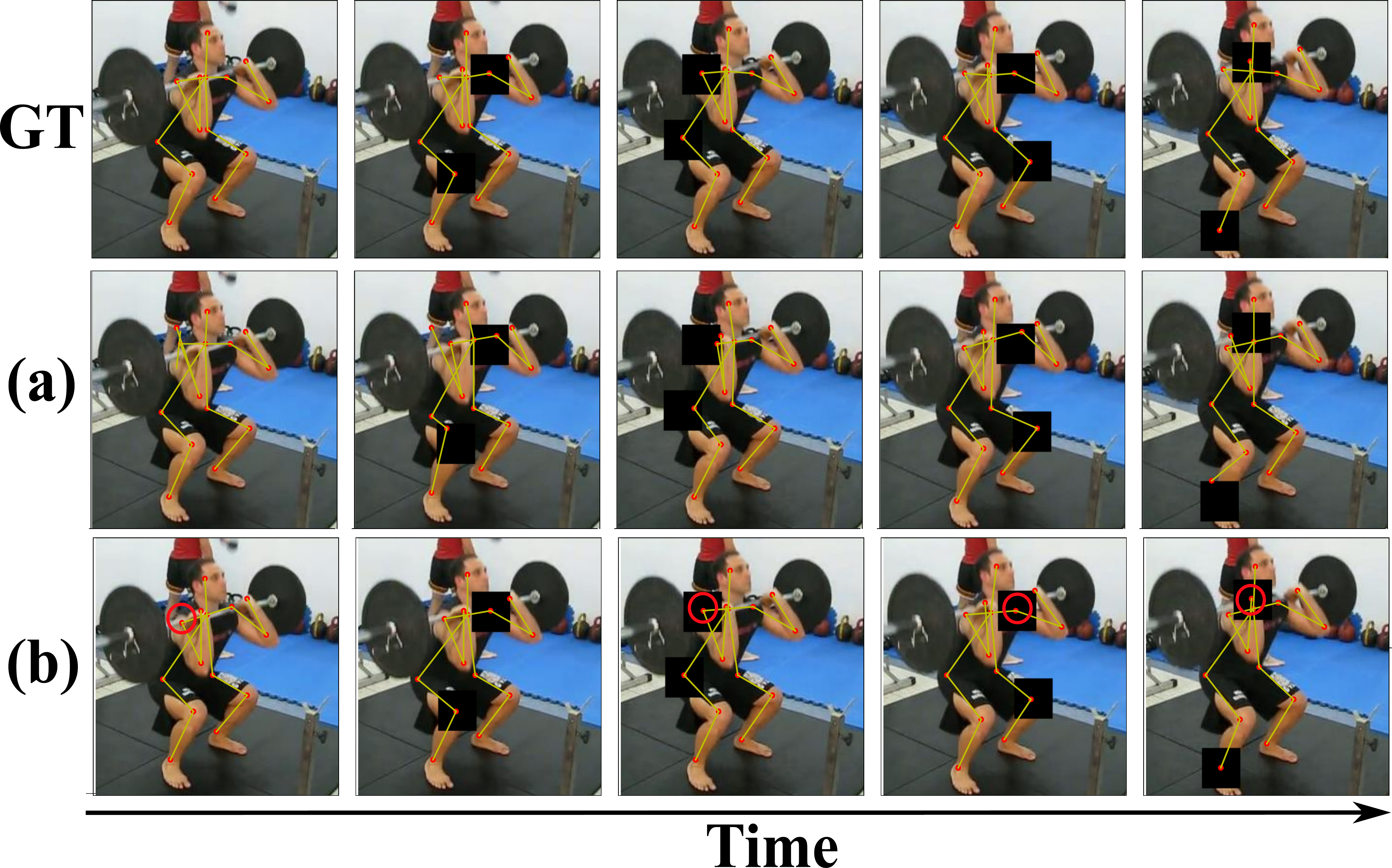}
        \label{fig:spatial_occlude_subfig_1}
    }
    \subfloat{
        \includegraphics[width=0.475\linewidth]{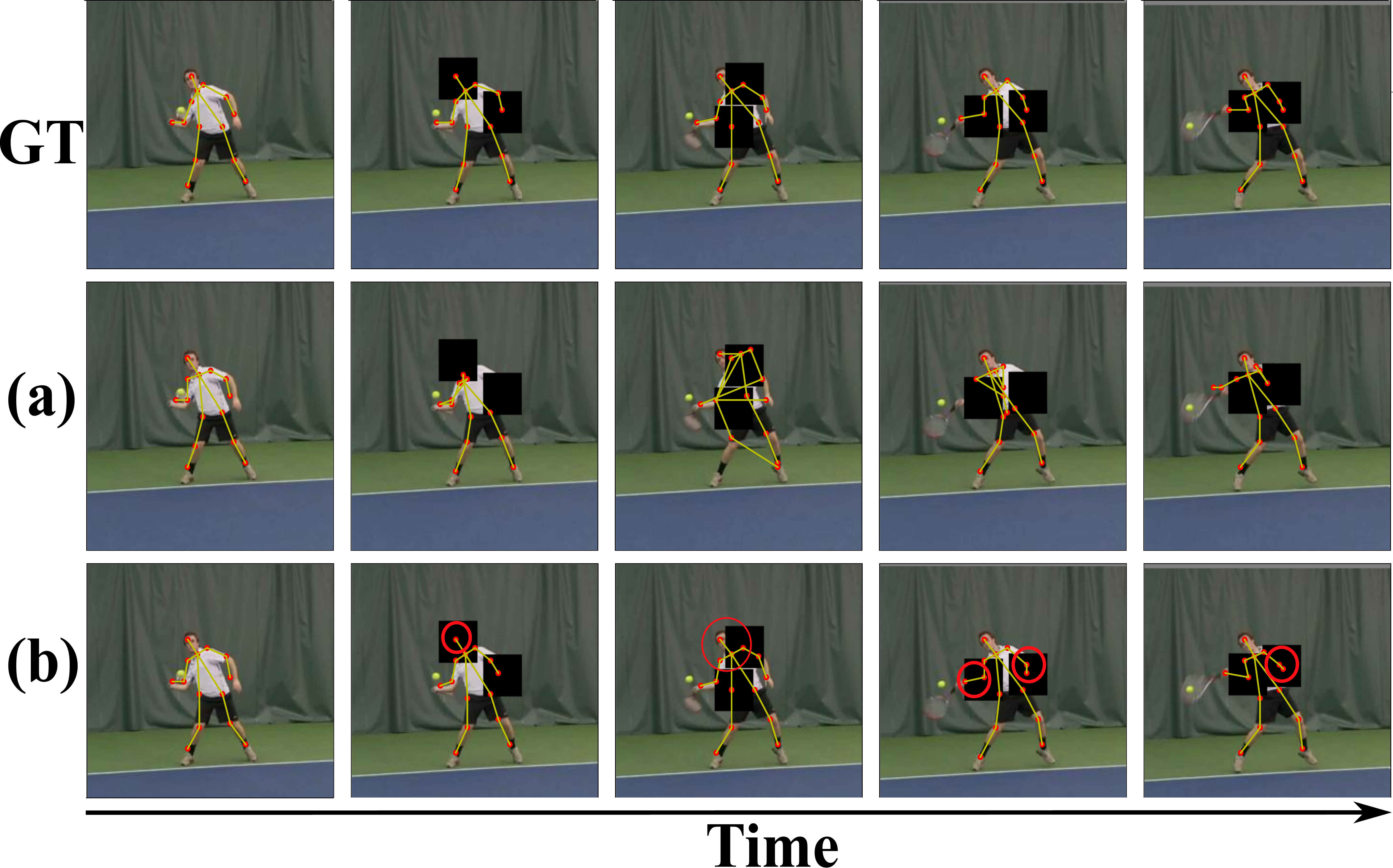}
        \label{fig:spatial_occlude_subfig_2}
    }%
   \caption{Visualization of occluded poses. (a) is the output of RPSTN \cite{RPSTN}. (b) is the output of KIMNet. We use red circles to highlight the joints that are localized more accurate by KIMNet.}
   \label{fig:occluded_pose}
\end{figure*}

 \textbf{Verification of spatial occlusion.} To explore the potential of KIMNet for occlusion, we randomly generate two masks from the $2^{nd}$ to the $T$-th frame, as shown in Figure\ref{fig:occluded_pose}. Because the spatial information of the occluded pose is missing, it is difficult for RPSTN to match similar regions in adjacent frames. Therefore, RPSTN provides inferior results for occluded joints. The proposed KIMNet can utilize all joints in the $1^{st}$ frame to locate occluded joints. It can provide reasonable positions for occluded joints in subsequent frames.
 
 % \begin{figure}[htbp]
 %     \centering
 %     \includegraphics[scale=0.11]{imgs/spatial_occlude_2.pdf}
 %     \caption{Visualization of occluded poses. We use red circles to highlight the joints that are localized more accurate by KIMNet.}
 %     \label{fig:occluded_pose}
 % \end{figure}
 
\begin{figure}[htbp]
     \centering
     \includegraphics[scale=0.6]{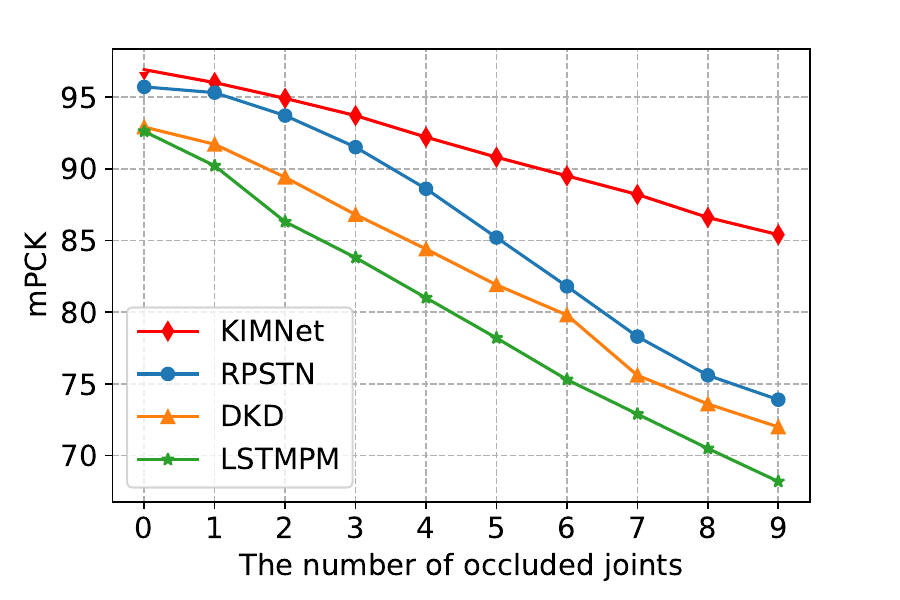}
     \caption{Comparison of the performance between KIMNet and existing methods under the occlusion scene.}
     \label{fig:occlude_curve}
 \end{figure}

 \textbf{Influence of the number of occluded joints.} To evaluate the influence of the number of occluded joints, the number of occluded joints is increased from 1 to 9. Under the same experimental settings, we compare the performance of the proposed KIMNet with existing methods, as shown in Figure\ref{fig:occlude_curve}. With the increase in the number of occluded joints, the performance of all models declines. However, compared to previous approaches, our KIMNet is more robust to occlusion. 
 
\subsubsection{The Naturally Occluded Scenario}
 
\begin{figure}[htbp]
  \centering
  \includegraphics[scale=0.09]{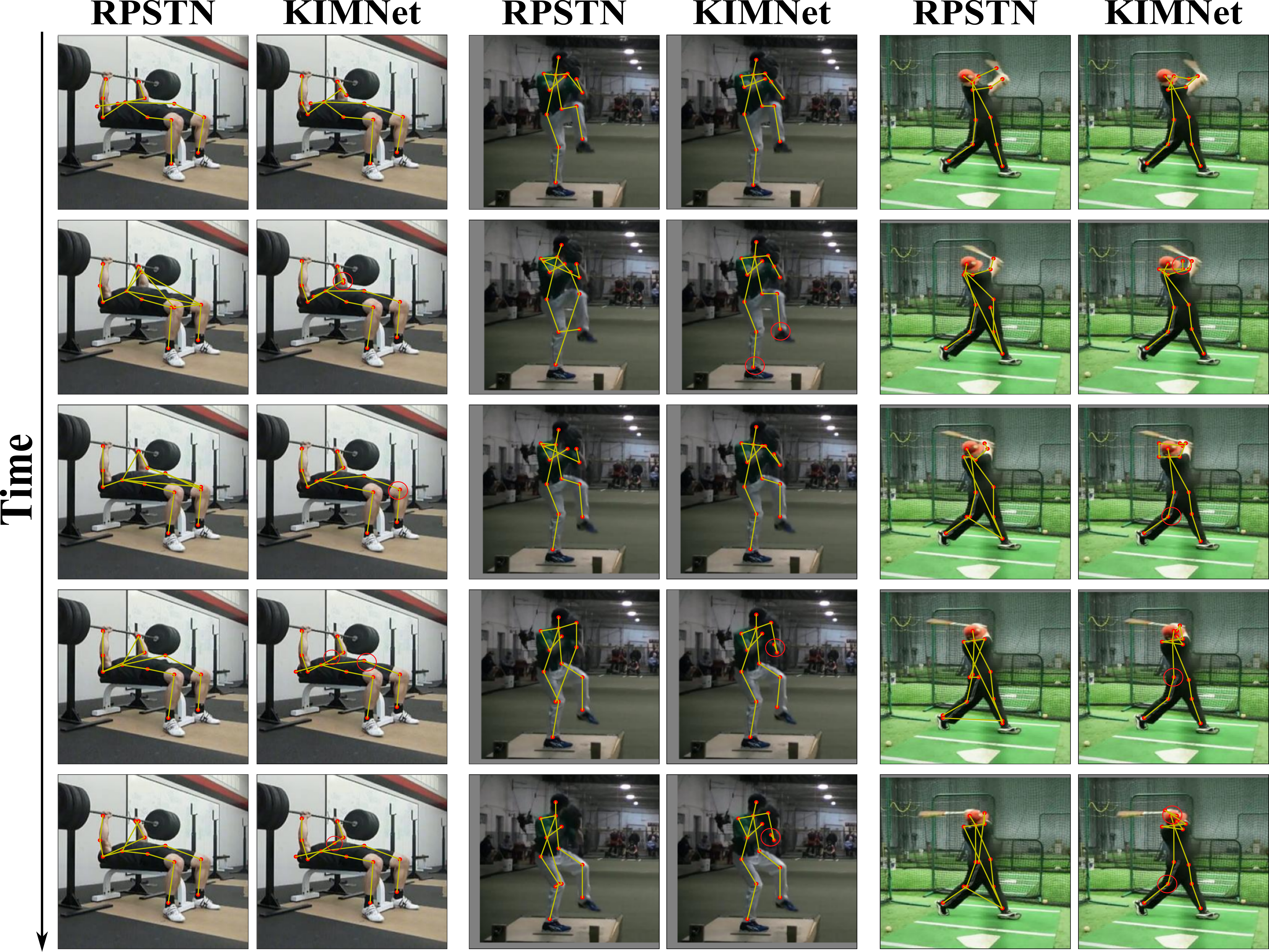}\\
  \caption{Experimental results in naturally occluded scenario. Joints that the KIMNet estimates better than the RPSTN are marked by the red circles.}
  \label{fig:natural_occ}
 \end{figure}

 To explore the potential of the KIMNet, we evaluate it on the naturally occluded scenario. Qualitative results are shown in Figure \ref{fig:natural_occ}. Because the KIMNet considers the temporal dependency between the joint in the current frame and all joints in the previous frame, the KIMNet is able to locate the current joint according to the information of all joints in the previous frame. Therefore, the proposed KIMNet provides more reasonable positions of the occluded joints (such as the joints in red circles).

\section{Conclusion}
\label{sec:conclusion}
Capturing temporal correlations between joints at different time is the crucial issue for video-based pose estimation. In this paper We propose a kinematics modeling network (KIMNet) that equipped with a plug-and-play kinematics modeling module (KMM) to capture each joint's motion cues by modeling the temporal correlation between any two joints across frames. In this way, the model can locate the current joint according to all joints' information in the previous frame, rather than only relying on the current joint's information. Especially, for the joints suffering from the occlusion or motion blur, information from other joints is beneficial for locating them. Experimental results show the advantages of the proposed KIMNet in achieving state-of-the-art results on two challenging benchmarks. %However, a drawback of our KIMNet is its long-term temporal modeling capability, which we aim to improve in future work.}

 \textbf{Limitation and Discussion.} Although the proposed KIMNet performs well against state-of-the-arts on the challenging datasets, it also suffers from some dilemmas in extremely complex scenarios, such as the severely occluded or truncated poses. For the heavily occluded poses, almost half of the joints are invisible, making it difficult to locate those invisible joints through several visible joints. In this case, one possible solution is that the symmetry of the human body can be used to help locate the joints on the occluded side of the body. For the severely truncated poses, the structural information of human poses is destroyed to a certain extent, which makes it challenging for the model to learn the correlation between joints. In this case, the possible solution is to introduce the prior knowledge of the human body to assist model learning. Therefore, we will do further research from above two aspects in future work.

\section*{Acknowledgments}
This work was supported partly by the National Natural Science Foundation of China (Grant No. 62173045, 61673192), partly by the Fundamental Research Funds for the Central Universities (Grant No. 2020XD-A04-3), and the Natural Science Foundation of Hainan Province (Grant No. 622RC675).

% \noindent
% \textbf{Acknowledgement:} This work is supported by NSF Data Science, Data Analytics that are Robust and Trusted
% (DART). 

% \newpage

\bibliographystyle{elsarticle-num}
\bibliography{egbib}

\clearpage
\clearpage

\end{document}